\documentclass{article}

\usepackage[accepted]{icml2026}

\makeatletter
\renewcommand{\ICML@appearing}{%
\textit{Accepted to the 2nd Workshop on Compositional Learning at ICML 2026,
Seoul, South Korea.
Copyright 2026 by the author(s).}}
\makeatother

\usepackage{amsmath}
\usepackage{amssymb}
\usepackage{booktabs}
\usepackage{multirow}
\usepackage{graphicx}
\usepackage{hyperref}
\usepackage{xurl}
\usepackage{xcolor}
\usepackage{microtype}
\usepackage{xspace}
\usepackage{enumitem}
\usepackage{rotating}
\usepackage{adjustbox}
\usepackage{array}
\usepackage{enumitem}

\usepackage{todonotes}

\newcommand{\icl}{ICL\xspace}
\newcommand{\xai}{XAI\xspace}
\newcommand{\dl}{DL\xspace}
\newcommand{\nle}{NLE\xspace}

\icmltitlerunning{%Explaining is Harder Than Predicting Alone: Towards a Formal Methodology for MLLMs as ICL Visual Classifiers
Explaining is Harder Than Predicting Alone: Evaluating Concept-based Explanations of MLLMs as ICL Visual Classifiers}
\begin{document}

\twocolumn[
%\icmltitle{Semantic Explainability %and Reasoning 
%Capabilities of
%  In-Context Learning multimodal {LLMs} %for Image Classification
  %as image classifiers
%  classifying images}
\icmltitle{Explaining is Harder Than Predicting Alone: Evaluating Concept-based Explanations of MLLMs as ICL Visual Classifiers}
% ---- Authors: LEAVE ANONYMOUS FOR SUBMISSION ----
 \begin{icmlauthorlist}
 \icmlauthor{Carmen Quiles-Ramírez}{UGR}
 \icmlauthor{Leticia L. Rodríguez}{UBA}
 \icmlauthor{Nicolás Martorell}{UBA}
  \icmlauthor{Natalia Díaz-Rodríguez}{UGR}
 \end{icmlauthorlist}
 % DESPUÉS:
 \icmlaffiliation{UGR}{Universidad de Granada, Spain}
 \icmlaffiliation{UBA}{Universidad de Buenos Aires, Argentina}
 \icmlcorrespondingauthor{Carmen Quiles-Ramírez}{carmenqr@ugr.es}

\icmlkeywords{In-Context Learning, Multimodal LLMs, Explainable AI, Concept-based Explanations, Description Logics, XAI Evaluation Metrics, Visual Classification}

\vskip 0.3in
]

\printAffiliationsAndNotice{}
%\cnat{googlear este error authorerr}
%\clet{lo investigue, era pq teniamos los autores comentados. No puede hacer que no salgan. Lei el sty y no encontre nada parece que funciona asi el template. Si alguno averigua algo más, avise. El error no sale más pero salen autores anonimos, no lo oculta.}

% ============================================================

\begin{abstract}
In-context learning (\icl) enables multimodal large language models (MLLMs) to classify images from a few labelled examples. Yet \emph{how} these models use the provided context remains opaque. 
While Chain-of-Thought prompting is widely used, recent work argues it may not reflect true internal computation~\citep{barez2025chain}. 
In this paper, we systematically evaluate the concept-based explainability of frozen MLLMs under few-shot \icl using five conditions of increasing formal rigour: from baseline classification to Description Logics (\dl) axiom generation. Evaluating four state-of-the-art MLLMs via an independent LLM-as-a-judge pipeline, we demonstrate that \emph{explaining is genuinely harder than predicting alone}. 
Surprisingly, forcing models to generate formally structured, concept-based explanations degrades predictive accuracy monotonically (from 93.8\% to 90.1\%), contradicting the assumption that explicit reasoning universally aids performance. 
However, when models successfully articulate class-discriminative visual features, explanation quality strongly correlates with correct predictions. Our findings suggest that while MLLMs excel at visual classification, they lack the specific instruction-tuning required for formal, machine-verifiable explainability.
\end{abstract}

% ============================================================
\section{Introduction}
\label{sec:intro}

In-context learning (\icl)~\citep{brown2020language} has emerged as a
powerful paradigm for adapting large language models to new tasks
without parameter updates.
Extended to the multimodal setting, MLLMs can now classify images from
a few labelled examples provided directly in the context
window~\citep{dong2024survey}.
Despite impressive few-shot accuracy, the reasoning process underlying
these predictions remains a black box: models produce answers without
any principled account of \emph{why} a particular image belongs to a
given class.

The dominant approach to eliciting explanations from large models is
chain-of-thought (CoT) prompting, which asks the model to verbalise
intermediate reasoning steps.
However, \citet{barez2025chain} argue that CoT does not constitute
genuine explainability: the generated text may diverge from the
model's actual internal computation, offering only a plausible-sounding
post-hoc narrative.
Similarly, \citet{huang2025mimicking} show that VLMs tend to mimic
rather than reason from \icl context, further questioning whether
CoT-style outputs constitute genuine understanding.
This gap motivates the search for \emph{concept-based explanations} \cite{poeta2025concept} ---formal
statements that are both human-readable and machine-verifiable.

We approach this problem through the lens of eXplainable AI
(\xai)~\citep{ali2023explainable} and Description
Logics~\citep{baader2009description}, a decidable fragment of
first-order logic widely used for knowledge representation.
By designing a hierarchy of five explanation conditions of increasing
formal complexity, we systematically investigate whether state-of-the-art
MLLMs can extract the discriminative visual features of an image class,
formalise them as IF--THEN rules, and ultimately express them as \dl.
We evaluate explanation quality using
an independent LLM-as-a-judge pipeline and a set of contributed XAI metrics.

Studying explanations within few-shot image classification 
provides a highly controlled setting. It encourages the model to ground its responses in newly introduced classes, helping to mitigate the influence of broad pre-training biases.
Our primary goal is to assess how MLLMs explain the knowledge derived from the prompt. We believe that narrowing the scope to visual data under the widely adopted few-shot paradigm provides a controlled and reproducible setting for studying concept-based explainability, as visual features are directly verifiable against the query image. 

\textbf{Contributions.} This paper makes the following empirical contributions:
\begin{itemize}[leftmargin=*, nosep]
  \item A systematic study of five explanation conditions of increasing
    formal rigour---from free-text \nle to \dl axiom generation---applied
    to few-shot image classification under \icl.
  \item A reproducible experimental protocol over four image
    classification datasets and four MLLMs, using fixed few-shot
    samples for direct cross-model and cross-condition comparability.
  \item An LLM-as-a-judge evaluation framework with nine explanation-quality metrics covering textual groundedness, hallucination free,
    concept counting, comprehensibility, conciseness, specificity, local discriminativeness, instruction following and logical coherence.
  \item Empirical evidence that concept-based explanation capability varies
    substantially across current MLLMs, and that classification accuracy
    alone does not reflect the quality of the generated explanations.
\end{itemize}

The rest of the paper is organized as follows: Section~\ref{sec:related} discusses related work. Section~\ref{sec:method} presents our methodology for assessing the concept-based explanation capabilities of MLLMs in visual classification tasks, formalizing the distinct types of explanations evaluated. Section~\ref{sec:evaluation} details the semantic explainability evaluation framework and introduces the judge model used to systematically assess these explanations. Section~\ref{sec:experiments} describes the experimental design, outlining the selected models, datasets, and overall setup. Section~\ref{sec:results} reports our experimental results, followed by a comprehensive discussion of these findings in Section~\ref{sec:discussion}. Finally, Section~\ref{sec:conclusion} concludes our study and highlights promising avenues for future work.

%\cnat{Si sobra espacio al final habria que separar las secciones del parrafo anterior e incluir al final de esta seccion un parrafo: The rest of the paper is organized as follows: Section x does y, section z does w..etc. En journals al menos se hace asi, verificar en algun ICML paper porque se ve un poco raro mezclando contribuciones con links a las secciones}

% ============================================================
\section{Related Work}
\label{sec:related}

\textbf{Few-shot In-Context Learning (Few-shot ICL).} 
It has emerged as a powerful paradigm for adapting Large Language Models to novel tasks using few-shot demonstrations, avoiding the need for weight updates \citep{brown2020language, dong2024survey}. Extensive research has focused on optimizing this technique and understanding its vulnerabilities, such as prompt sensitivity and the impact of demonstration quality \citep{zhao2021calibrate, lu2022fantastically}. A critical ongoing debate surrounding ICL questions the extent to which models truly learn from the provided context versus merely retrieving prior knowledge from their pre-training phase \citep{min2022rethinking, webson-pavlick-2022-prompt}. Recently, this paradigm has been extended to the visual domain, where works such as \citet{zhang2023makes}, \citet{zhou2024visual}, and \citet{sun2025exploring} explore strategies to improve multimodal ICL performance. However, these studies evaluate success almost exclusively by classification accuracy. %Our work focuses on explainability, investigating whether Multimodal Large Language Models can produce structured and faithful explanations to justify their in-context behavior.
Crucially, in all \icl settings --- including zero-, one-, and few-shot --- the model's weights remain frozen; no gradient update occurs~\citep{brown2020language}. Our work operates in this paradigm, investigating whether frozen MLLMs can produce concept-based explanations to justify their in-context behaviour.

\textbf{Explainability of language models.}
Natural language explanations (NLEs) for model predictions have been
studied as a route to interpretability~\citep{wang2025multimodal}.
\citet{barez2025chain} provide a critical analysis showing that CoT
traces do not reliably reflect internal computation.
Furthermore, previous authors warn of the safety risks associated with relying on CoT explanations, demonstrating that models frequently ignore their own reasoning traces during inference \cite{lanham2023measuring} and often produce plausible but misleading rationales \cite{turpin2023language}
Probing studies have further shown that pre-trained language models
specialise at different aspects of language understanding, with
performance varying substantially across linguistic
tasks~\citep{gjinika2024analysis}.
The broader landscape of XAI in the era of LLMs is surveyed
in~\citet{polignano2024xai}, which highlights the growing need for
systematic evaluation frameworks.
We take this body of work as a starting point and focus specifically
on a gap it does not address: the capacity of MLLMs to \emph{explain}
image classification decisions under few-shot \icl---as opposed to
merely producing accurate labels.

\textbf{Structured and neurosymbolic explanations.}
Description Logics~\citep{baader2009description} provide a decidable,
ontology-compatible formalism for expressing class definitions as
verifiable axioms.
Recent neurosymbolic work has explored symbol
grounding~\citep{ontiveros2025grounding, marconato2025symbol} and
logic-explained networks~\citep{ciravegna2023logic}, but these
approaches require supervised training.
Our contribution is complementary to these supervised approaches:
rather than training models to produce formally verifiable axioms, we
probe whether frozen MLLMs can spontaneously generate \dl %\ccarm{Aqui pongo DL inspired porque no generamos salidas OWL}\cnat{ no se puede usar eso, es DL exacto o no lo es, o lo hace bien el modelo o mal, no podemos controlarlo, nunca decir dl inspired}
concept-based explanations at test time---without any fine-tuning---and
evaluate their quality through an independent judge pipeline.

\textbf{The cost of explicit reasoning.}
While explanations are desirable for transparency, recent work suggests that forcing models to articulate intermediate steps can sometimes degrade performance. \citet{liu2025mindyourstep} demonstrated that Chain-of-Thought reasoning can reduce accuracy on tasks where human performance also drops when forced to think step-by-step, as it biases the model toward over-generalised rules rather than explicit context. Furthermore, the interplay between in-context examples and pre-trained knowledge remains complex, with models often struggling to suppress strong pre-training priors even when provided with explicit demonstrations \citep{zhang2024mystery}. This highlights a potential conflict between optimizing for predictive accuracy versus explanatory transparency under \icl.

% ============================================================
\section{Method: Assessing Concept-based Explanation Capabilities of MLLMs for Image Classification}
\label{sec:method}

In this section, we outline our experimental methodology. We first define the formal few-shot in-context learning setup for image classification, and subsequently detail the five explanation conditions designed to systematically probe the models' reasoning and explainability capabilities.

\subsection{Problem Setup and methodology}
\label{sec:setup}

We frame the task as $N$-way, few-shot ($K$-shot) image classification under \icl.
Given a support set $\mathcal{S} = \{(x_i, y_i)\}_{i=1}^{N \times K}$
with $N$ classes and $K$ labelled examples per class, and a query image
$x_q$, the model must predict $\hat{y}_q \in \mathcal{Y}$ using only
the information provided in the context window---no fine-tuning or weight
updates are allowed~\citep{wang2020fsl, vinyals2016matching}.
We extend this standard setup by additionally requiring the model to
produce a structured \emph{explanation} of its prediction according to
one of five explanation conditions (Section~\ref{sec:conditions}).

\subsection{Evaluating five kinds of concept-based explanations with incremental semantic complexity. }
\label{sec:conditions}
%\cnat{para el llm son condiciones pero para el resto del mundo son tipos o formatos de explicacion. añadir siempre el guion a p3 en adelante: feature based-explanations, falta adjetivo en titulo de P3, p4. y p1 no es un baseline usar mi ultima correccion en el word: Classification with no explanation, puede dejarse el baseline si pones 2 Baseline: Classification with no explanation. Mejor ase. De donde viene la P si son explanation types o condiciones? nomenclatura hacerla coherente e intuitiva, si no nos referimos a estos Ps nunca mas, no añadir notacion innecesaria, numeros itemize enumerate basta o para ahorrar espacio en bold como esta ahora mejor 1.- 2. lo q sea. etc}
We design five explanation  types of increasing formal complexity to evaluate the explanation capabilities of multimodal LLMs.
All conditions requested share a common system prompt with three constraints
enforced via structured output formatting: (i)~predictions must be
grounded exclusively in visually observable evidence from the provided
examples; (ii)~external world knowledge, hidden assumptions, and
speculative attributes are explicitly prohibited; (iii)~the final class
label must appear inside a \texttt{<response>} XML tag, whose content
must be copied verbatim from the provided options list, to enable
deterministic parsing. The common theme across all five conditions is that they are centered around concept-based explainability~\citep{poeta2025concept}: unlike saliency maps or gradient-based methods, all explanation types rely on human-understandable visual concepts. The exact system prompts and formatting constraints used across all explanation conditions are detailed in Appendix~\ref{app:prompts}.

\begin{description}[leftmargin=0pt]
  \item[\textbf{E1 --- Classification with no explanation (Baseline).}]
    We request the model to output only the final class label inside
    \texttt{<response>}.
    No explanation nor intermediate output is requested.
    This establishes the accuracy baseline against which all other
    explanation conditions are compared.

  \item[\textbf{E2 --- Natural Language Explanation (\nle).}]
    We ask the model to produce a concise natural language explanation inside an
    \texttt{<explanation>} tag, grounded only in observable visual
    evidence, briefly justifying why that evidence supports the selected class over the alternatives.
    This corresponds to standard CoT-style prompting. %\cnat{ quizas un footnote especificando el thinking option o similar empleado para enforzar este setting en cada modelo? sino en apendix aclarar para cada uno que puedan reproducirlo} SE AGREGO EN OTRA PARTE

  \item[\textbf{E3 --- Features-based explanation.}]
    We prompt the model to identify the minimal sufficient set of critical, concrete, observable visual features needed to support the classification, expressed as short noun phrases in bullet-point form inside a \texttt{<features>} tag.
    %\cnat{ pusimos bullet points? creo q no}inside a \texttt{<features> -respuesta: lo devuelve en Bullet} 
    %Abstract interpretations, causal explanations \cnat{ de hecho fomentamos estas en una metrica,distinctiveness creo, nunca las evitamos! revisar que paso aqui, no contradecir los prompts!}, and full sentences
    %are explicitly disallowed.

  \item[\textbf{E4 --- Feature-value pairs explanation.}]
    We ask the model to list the observable features of the query image in
    a \texttt{<features>} tag, then to construct a knowledge base of
    IF--THEN rules grounded in visual patterns from the labelled
    examples in a \texttt{<kb>} tag, and finally
    we ask it to identify which rule is best satisfied by the query image in a
    \texttt{<rule\_check>} tag---referring only to the features already listed and introducing no new evidence.
    This tests whether the model can perform structured logical reasoning from few-shot examples.

  \item[\textbf{E5 --- DL Axioms explanation.}]
    Description Logics (\dl)~\citep{baader2009description} is a decidable fragment of first-order logic designed for knowledge representation.
    A \emph{knowledge base} $\mathcal{KB} = (\mathcal{T}, \mathcal{A})$
    consists of a TBox $\mathcal{T}$ (terminological axioms defining concept hierarchies and roles) and an ABox $\mathcal{A}$ (assertional
    facts about individuals).
    DL reasoning supports inference operations including subsumption, consistency checking, and instantiation, making axiom-based
    explanations formally verifiable. %\carmen{Importante que verifique Natalia}
    
    We ask the model to present %\cnat{me parece demasiado asertivo, the model produces, the    model represents, los comienzos d estas frases de Pi. Cambiar todos estos Ps a The          model is evaluated on producing.... (no The model produces....porque es lo q le             pedimos, no lo que hace) en todos los Ps} 
    %\cleti{lo cambio}
    the classification evidence using ontological
        \dl statements~\citep{baader2009description}: a
        \texttt{<tbox>} tag defines formal class conditions as necessary,
        sufficient, or necessary-and-sufficient axioms using
        \texttt{hasVisualFeature} roles derived from the labelled examples;
        an \texttt{<abox>} tag records the specific property assertions
        observed in the query image; and a \texttt{<dl\_explanation>} tag
        explains how the ABox assertions satisfy the TBox axioms to
        logically derive the predicted class.
        This is the most constrained and formally rigorous explanation
        condition.
    \end{description}    

% ============================================================
\section{Semantic explainability evaluation}
\label{sec:evaluation}
\begin{figure*}[htbp!]
    \centering
    \includegraphics[width=1\linewidth]{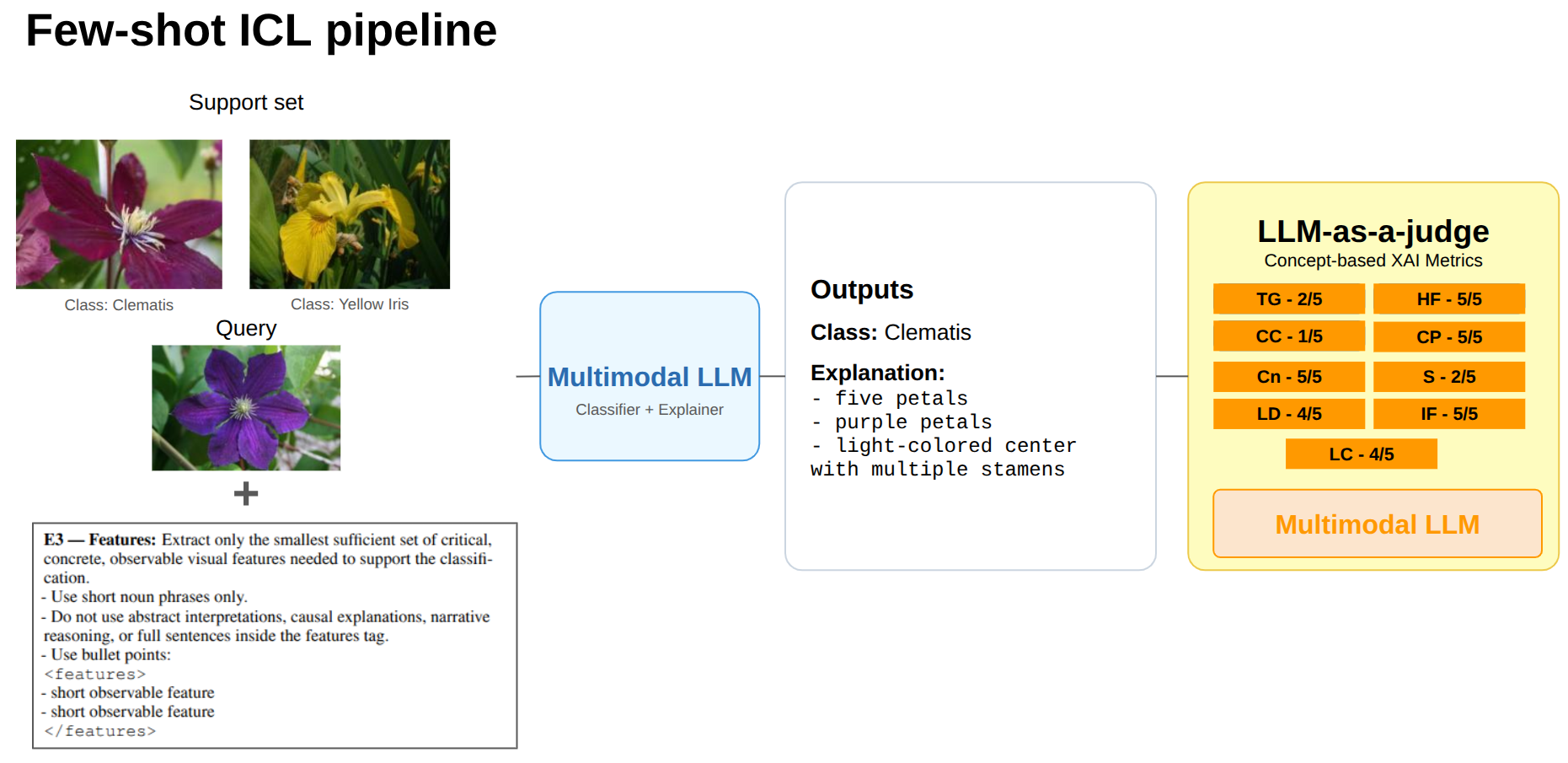}
    \caption{Overview of the In-Context Learning \textit{Explain-Classify and Evaluate} pipeline. The Multimodal LLM acts as both classifier and explainer, while a separate LLM-as-a-judge evaluates nine concept-based XAI metrics (1--5\,$\uparrow$): see Table~\ref{tab:metrics}. The example shown corresponds to E3 (feature-based explanation), where the model is prompted to identify the minimal sufficient set of critical, concrete, and observable visual features needed to support the classification. The judge, using a different MLLM and prompt, scores the generated explanation across the same nine evaluation dimensions.}
\label{fig:pipeline}
\end{figure*}
%\cleti{Cambio la introducción dado que explicaba accuracy. Es un termino muy frecuente, no hay que indicar como se calcula. Si quisieramos explicarlo, deberia ser una formula. Es un concepto muy básico para AI, lo dejaría sobrentendido.}
%\ccarm{Cambiar a gpt5 mini} %\cleti{Muevo lo de gpt5 a la parte de experimentacion}
We use accuracy to measure the quality of classified instances. Due to the fact that the classification accuracy metric cannot capture the \emph{quality} of the generated explanations, we use an independent \textit{LLM-as-a-judge} pipeline (Figure \ref{fig:pipeline}). %\cnat{ quizas aqui citar alguno o el que nos inspiro para nuestra rubrica, podemos decir inspirado en ... y concretizando que usamos un llm diferente al classifier and explainer LLM?en el LIAA mandaron uno bueno de como evaluar, habria que ver si seguimos su advice y si si, citarlo, future work sino}. 
Our approach uses a single answer that scores 9 different XAI metrics, as shown in  Table \ref{tab:metrics}. This enables scalable benchmarking and fast iterations at the scale of the full experiment; the alignment of automated scores
with human judgement is validated in Appendix~\ref{app:human_validation}. This work proposes a standardized way of evaluating the explanations.

\begin{table}[htbp!]
\centering
\caption{Explanation quality metrics we define to be scored by the LLM-as-a-judge
  (1\,=\,worst, 5\,=\,best). The complete scoring rubrics for each proposed metric are detailed in Appendix~\ref{app:metrics}.}
\label{tab:metrics}
\small
\begin{tabular}{@{}lp{4cm}@{}}
\toprule
\textbf{Concept-based XAI Metric} & \textbf{Description} \\
\midrule
Textual Groundedness (TG)   & All salient image concepts are mentioned in the explanation.
\\
Hallucination Free (HF)     & Every claim is visually verifiable in the image. \\
Concept Counting (CC)      & %Quantitative counts is redundant y no dice nada
Concept counts match the image exactly. \\
Comprehensibility (CP)     & Explanation readable, accessible to non-expert users.\\
Conciseness (Cn)           & Conveys only necessary information without redundancy. \\
Specificity (S)           & Relies on precise, non-generic details of the sample. \\
Local Discriminativeness (LD) & Highlights features distinguishing %the 
predicted class from others. \\
Instruction Following (IF)  & Complies with all explicit format and structure constraints. \\
Logical Coherence (LC)     & Sentences build into a smooth, valid deduction. \\
\bottomrule
\end{tabular}
\end{table}

For each query trial the judge receives: (i)~the query image,
(ii)~the set of candidate class labels, (iii)~the classifier's full
output including the predicted class label, (iv)~a description of the
explanation condition being evaluated (E1--E5), and
(v)~the scoring rubric.

The judge operates in a separate pipeline only receiving the detailed
information and in a zero-shot setting: it does not receive the
few-shot support images seen by the classifier model, due to compute
constraints. Evaluation is grounded exclusively in the query image
as the visual source of truth. For the one metric that relies on the
judge understanding attributes about the set of candidate classes
(e.g.\ Local Discriminativeness), we expect the judge model's prior
knowledge to be sufficient within the selected datasets and examples.

To optimise this expectation while remaining below compute budget
limits, we choose \texttt{gpt-5-thinking-mini}~\citep{openai2025gpt5}
as the judge model, a medium-size state-of-the-art reasoning model,
accessed via the OpenRouter API with \texttt{reasoning\_effort} set
to \texttt{medium}.
To assess whether judge access to support images affects the key
Local Discriminativeness metric, we conduct a controlled ablation
study using a second judge model;
see Appendix~\ref{app:judge_validation}.

To further validate the judge pipeline, we conducted an independent
human evaluation in which three researchers manually scored a
stratified sample of 192 trials using the same 1--5 rubrics
(see Appendix~\ref{app:human_validation}).
The judge demonstrates strong alignment with human judgements:
Spearman $\rho > 0.60$ against the human mean across 6 of 9 metrics,
including Local Discriminativeness ($\rho = 0.656$), the metric
most central to our findings, confirming the validity of the
automated evaluation pipeline.

% To optimise this expectation while remaining below compute budget
% limits, we choose \texttt{gpt-5-thinking-mini}~\citep{openai2025gpt5}
% as the judge model, a medium-size state-of-the-art reasoning model,
% accessed via the OpenRouter API with \texttt{reasoning\_effort} set
% to \texttt{medium}.
%\ccarm{Si finalmente se queda asi argumentar por que no pasamos las support images} %Todo FUture work to do indeed

% ============================================================
\section{Experiments}
\label{sec:experiments}
%\cnat{no puede haber secciones vacias como ahora la 5, no se empieza nunca con 5.1 directamtne. resumir resto de la secion lo mas rapido posible, q ocupe poco siempre una frase al menos para introducir rest of sections}. %\cnat{usar comando [htbp!] para que las tablas queden donde se pone en el texto y como la 8 no se metan entre las referencias.}
In this section, we present our experimental setup.
%\cnat{no puede ir la tabla antes de introducirla}

\textbf{Datasets and Models}
\label{sec:datasetsandmodels}

Following the standard VLM few-shot benchmark established by
\citet{zhou2022learning} and widely adopted in the VLM literature, we
used four datasets that cover a diverse range of visual
recognition tasks:
CIFAR-10~\citep{krizhevsky2009cifar} (10 generic object
categories, low-resolution images),
DTD~\citep{cimpoi2014dtd} (47 texture classes),
Oxford Flowers~102~\citep{nilsback2008flowers}
(102 flower species), and
Oxford-IIIT Pets~\citep{parkhi2012pets}
(37 fine-grained pet breed categories).
Beyond their role as a standard benchmark, these datasets are
particularly well-suited for evaluating the quality of textual
explanations: Oxford-IIIT Pets and Flowers~102 involve classes that
are highly visually describable, providing rich discriminative
features for the model to articulate; DTD requires describing abstract
perceptual properties such as texture and material, testing whether
models can express non-object concepts; and CIFAR-10 serves as a
coarser-grained contrast baseline.

We evaluated four state-of-the-art MLLMs selected to represent
a diverse range of model families, scales, and access types
(Table~\ref{tab:models}).
All models support native multi-image input, a requirement for the
$N$-way $K$-shot context format used in our experiments,
and are instruction-tuned~\citep{wei2021finetuned, ouyang2022training}.
All are accessed via the OpenRouter API at temperature $T{=}0$.
Model selection was further guided by recent systematic benchmarks of
multimodal foundation models~\citep{ramachandran2025evaluating}
and multimodal \icl tasks~\citep{huang2025mimicking}

\begin{table}[htbp!]
\centering
\caption{MLLMs evaluated in this study. $\dagger$~17B active (MoE). $\ddagger$~4B active (MoE).
  CTX = context window. See Appendix~\ref{app:models} for details.}
\label{tab:models}
\setlength{\tabcolsep}{3pt}
\begin{tabular}{@{}lc>{\centering\arraybackslash}p{0.15\linewidth}>{\raggedleft\arraybackslash}p{0.15\linewidth}@{}}
\toprule
\textbf{Model} & \textbf{Type} & \textbf{Params} & \textbf{CTX} \\
\midrule
Gemini 2.5 Flash  & Proprietary      & ---                & 1{,}048K \\
Gemma 4 26B             & Open (MoE) & 26B$^{\ddagger}$   & 262K \\
Qwen3 VL 8B            & Open       & 8.77B              & 131K \\
LLaMA 4 Scout       & Open (MoE) & 109B$^{\dagger}$   & 328K \\
\bottomrule
\end{tabular}
\end{table}

\textbf{Experimental Setup}
\label{sec:setup_exp}
%\cnat{evitar tantos subtitulos para ahorrar espacio, la sec 5 se puede llamar como esta. Los parrafos de cada modelo pueden ir al apendix. No hace falta una subseccion como la actual 6.1.}

We generate support sets following the standard episodic few-shot
evaluation protocol~\citep{vinyals2016matching, snell2017prototypical}.
Each support set of $N$ classes with $K$ labelled images per class, and a single query image ($Q{=}1$).

Setting $Q{=}1$ serves two purposes.
First, it ensures that each trial constitutes a fully independent
sample---with its own support set---satisfying the independence
assumption required by the non-parametric statistical tests used in
our analysis (McNemar, Wilcoxon signed-rank, Friedman).
Increasing $Q$ within a fixed support set would yield pseudo-replicates
conditioned on the same context, artificially reducing variance
estimates and inflating statistical power.
Second, since our primary interest lies in the explanations generated
under each explanation condition, $Q{=}1$ ensures that each explanation
is produced independently, without influence from prior model outputs
within the same support set---critical for explanation conditions that
elicit structured reasoning (E4, E5), where accumulated context could
introduce consistency biases unrelated to the condition.

We evaluated all combinations of $N \in \{2, 3, 4\}$ and
$K \in \{1, 5\}$, following the canonical 1-shot and 5-shot evaluation
protocol established in the few-shot learning
literature~\citep{vinyals2016matching, snell2017prototypical}.
The 1-shot setting tests whether a single labelled example per class is
sufficient for the model to both classify and explain; the 5-shot setting
tests whether additional context improves accuracy and, crucially,
explanation quality.
The range $N \in \{2, 3, 4\}$ spans binary discrimination (N=2) through
increasingly fine-grained multi-class settings (N=4), allowing us to
assess how explanation quality scales with classification difficulty.
The number of independent support sets repetitions per configuration is
chosen such that $\mathrm{Reps} \times N = 12$ for all values of $N$,
ensuring that each difficulty level contributes an equal number of class
presentations to the aggregate evaluation.
Without this constraint, configurations with smaller $N$ would yield
more trials in total, inflating performance estimates for easier
settings and confounding comparisons across difficulty levels.
All support sets are generated once with a fixed random seed (42) and
reused across all models and explanation conditions.
Table~\ref{tab:setup} summarises the full configuration grid.

% Table 1 from PDF
%\cnat{ centrar todas las tablas con \centering antes de caption y despues del begin}
\begin{table}[htbp!]
\centering
\caption{Few-shot \icl configurations evaluated. Reps %\cnat{ ? repetitions? porque tienen que ser igual a 12, o que hay que balancear y por que? no se entiende \ccarm{Explciado justo en el parrafo arriba de la tabla}. Should Test be called Trial for consistency? intentar llamar siempre igual a las cosas}
balanced so that
  Reps~$\times$~$N = 12$. Total: $26 \times 4~\text{models} \times
  4~\text{datasets} \times 5~\text{conditions} = 2{,}080$ runs.}
\label{tab:setup}
\small
\setlength{\tabcolsep}{4pt}
\renewcommand{\arraystretch}{1.1}
\begin{tabular}{@{}ccccrr@{}}
\toprule
\textbf{Trial} & $N$ & $K$ & $Q$ & \textbf{Runs} &
  \textbf{Total runs / Trial} \\
\midrule
1 & 2 & 1 & 1 & 6 & 480 \\
2 & 2 & 5 & 1 & 6 & 480 \\
3 & 3 & 1 & 1 & 4 & 320 \\
4 & 3 & 5 & 1 & 4 & 320 \\
5 & 4 & 1 & 1 & 3 & 240 \\
6 & 4 & 5 & 1 & 3 & 240 \\
\midrule
Total & --- & --- & 1 & 26 & 2{,}080 \\
\bottomrule
\end{tabular}
\end{table}

% ============================================================
\section{Results}
\label{sec:results}

\textbf{Classification Accuracy.}
Tables~\ref{tab:acc_cond_model} and~\ref{tab:acc_cond_dataset} report mean accuracy aggregated over all models and datasets respectively. Overall accuracy across all conditions and models reaches 92.6\%. Gemini 2.5 Flash is the top-performing model in every condition (Table~\ref{tab:acc_cond_model}), and Flowers is the easiest dataset, with all models achieving near-perfect accuracy regardless of condition (Table~\ref{tab:acc_cond_dataset}). Critically, accuracy decreases monotonically as explanation complexity increases: from 93.8\% at E1 (Classification) to 90.1\% at E5 (DL Axioms), confirming that demanding richer explanations imposes a cost on predictive performance. The accuracy gap between conditions is most pronounced for Qwen3 VL 8B, which drops from 95.1\% at E1 to 83.0\% at E5, while Gemini 2.5 Flash remains above 95\% across all conditions. Dataset difficulty also varies: DTD shows the steepest decline across conditions (from 85.8\% at E1 to 82.6\% at E5), consistent with its abstract textural content being harder to verbalise formally. Detailed per-model and per-configuration breakdowns are provided in Appendix~\ref{app:extra}.

\begin{table}[htbp!]
\centering
\caption{Mean accuracy (\%) by explanation condition and model,
  aggregated over 4~datasets and 6~few-shot configurations
  ($N\in\{2,3,4\}$, $K\in\{1,5\}$): $4\times26=104$ observations per cell.
  $^\dagger$Classification-only baseline. Values: mean~($\pm$SE). Best per column in bold.} %\cnat{completar bolds? tb si en el texto nos referimos a Ei, mencionar las Ei en tablas 4 y 5 antes de la condicion}
\label{tab:acc_cond_model}
\small
\setlength{\tabcolsep}{3pt}
\renewcommand{\arraystretch}{1.1}
\begin{tabular}{@{}lcccc@{}}
\toprule
\textbf{XAI Condition}
  & \textbf{Gem.~2.5F} & \textbf{Gem.~4} & \textbf{Qwen} & \textbf{LLaMA}\\
\midrule
E1 - Classification$^\dagger$
  & 96.9~{\tiny(1.0)} & \textbf{94.4}~{\tiny(1.3)} & \textbf{95.1}~{\tiny(1.3)} & 88.5~{\tiny(1.9)} \\
E2 - NLE
  & \textbf{97.2}~{\tiny(1.0)} & 94.1~{\tiny(1.4)} & 92.7~{\tiny(1.5)} & \textbf{90.3}~{\tiny(1.7)} \\
E3 - Features
  & 96.9~{\tiny(1.0)} & 93.1~{\tiny(1.5)} & 93.8~{\tiny(1.4)} & 88.5~{\tiny(1.9)} \\
E4 - Feature-value pairs
  & 95.8~{\tiny(1.2)} & \textbf{94.4}~{\tiny(1.3)} & 92.4~{\tiny(1.6)} & 86.5~{\tiny(2.0)} \\
E5 - DL Axioms
  & 96.2~{\tiny(1.1)} & 92.4~{\tiny(1.6)} & 83.0~{\tiny(2.2)} & 88.9~{\tiny(1.9)} \\
\bottomrule
\end{tabular}
\end{table}

%\cnat{ no cabe el +- en el parentesis que se lea mejor de las STd dev?. No entiendo las negritas en tabla 4 y 5, no debe haber una negrita por cada mejor valor por cada condicion ? porque hay varios para feature value en tabla 4? sobra uno? no deberia haber una nerita por dataset en tabla 5? pues son datasets diferentes? (es decir una negrita para el mejor valor por columna?)}
\begin{table}[htbp!]
\centering
\caption{Mean accuracy (\%) by explanation condition and dataset,
  aggregated over 4~models and 6~few-shot configurations:
  $4\times26=104$ observations per cell.
  $^\dagger$Classification-only baseline. Values: mean~($\pm$SE).
  Best per column in bold.}
\label{tab:acc_cond_dataset}
\small
\setlength{\tabcolsep}{3pt}
\renewcommand{\arraystretch}{1.1}
\begin{tabular}{@{}lcccc@{}}
\toprule
\textbf{XAI Condition}
  & \textbf{Flowers} & \textbf{Pets} & \textbf{CIFAR} & \textbf{DTD} \\
\midrule
E1 - Classification% (baseline, no explanation)
$^\dagger$
  & \textbf{100.0}~{\tiny(0.0)} & \textbf{95.1}~{\tiny(1.3)} & 94.1~{\tiny(1.4)} & \textbf{85.8}~{\tiny(2.1)} \\
E2 - NLE
  & \textbf{100.0}~{\tiny(0.0)} & 93.8~{\tiny(1.4)} & \textbf{94.8}~{\tiny(1.3)} & \textbf{85.8}~{\tiny(2.1)} \\
E3 - Features
  & \textbf{100.0}~{\tiny(0.0)} & 93.4~{\tiny(1.5)} & \textbf{94.8}~{\tiny(1.3)} & 84.0~{\tiny(2.2)} \\
E4 - Feature-value pairs
  & 99.7~{\tiny(0.3)} & 93.1~{\tiny(1.5)} & 92.4~{\tiny(1.6)} & 84.0~{\tiny(2.2)} \\
E5 - DL Axioms
  & 96.5~{\tiny(1.1)} & 91.0~{\tiny(1.7)} & 90.3~{\tiny(1.7)} & 82.6~{\tiny(2.2)} \\
\bottomrule
\end{tabular}
\end{table}

\textbf{Effect of $K$ and $N$ on explanation quality.} Increasing support images from $K{=}1$ to $K{=}5$ improves accuracy by 7.0 pp ($p = 2.0 \times 10^{-13}$), with the largest explanation quality gain concentrated in Local Discriminativeness ($\Delta{=}{+}0.26$, $p_{\mathrm{Bonf}} = 1.5 \times 10^{-6}$). Conversely, increasing $N$ degrades accuracy monotonically and, among all metrics, only Local Discriminativeness falls significantly ($3.86$ at $N{=}2$ to $3.40$ at $N{=}4$; $p_{\mathrm{Bonf}} = 1.9 \times 10^{-10}$), confirming that adding classes stresses discriminative reasoning specifically. Results are detailed in Appendix~\ref{app:nk_analysis} (Figure~\ref{fig:nk_analysis}).

\begin{figure}[htbp!]
\centering
\includegraphics[width=0.75\linewidth]{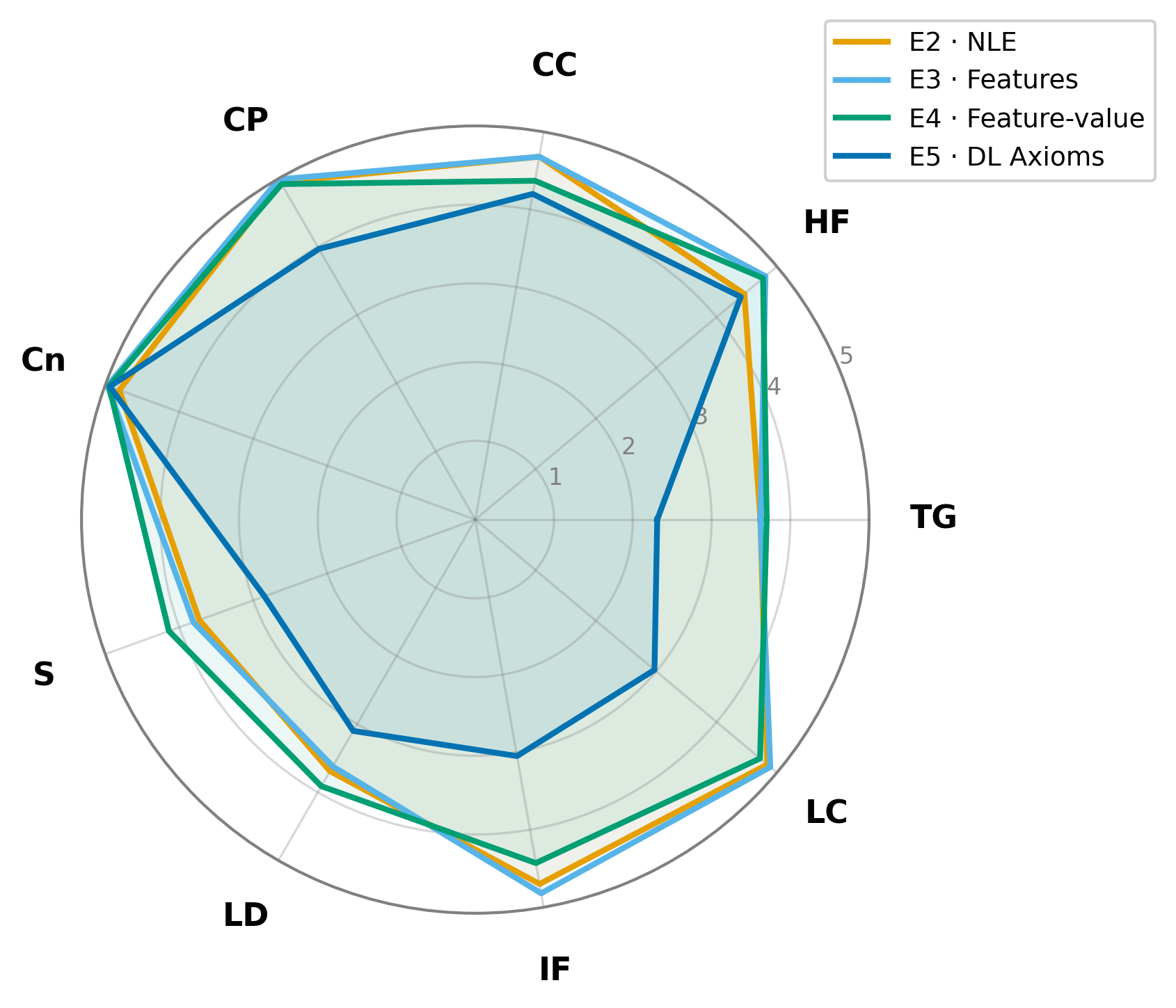}
\caption{LLM-as-a-judge scores across nine explanation quality metrics
  by condition (E2--E5), averaged over all models and datasets.
  DL Axioms (E5) shows a markedly smaller polygon, particularly
  on Textual Groundedness, Specificity, Instruction Following,
  and Logical Coherence. Metric abbreviations as in Table~\ref{tab:metrics}.}
\label{fig:radar_prompt}
\end{figure}

\textbf{Explanation Quality across explanation conditions.}
Table~\ref{tab:judge_cond_metric} reports mean LLM-as-a-judge scores across the nine quality metrics. DL  Axioms (E5) scores substantially lower than other conditions on Textual Groundedness (2.31), Specificity (2.85), Local Discriminativeness (3.10), Instruction Following (3.05), and Logical Coherence (2.97), while performing comparably on Hallucination Free (4.40) and Conciseness (4.94). This pattern is clearly visible in Figure~\ref{fig:radar_prompt}: DL Axioms produces a markedly smaller polygon than the other conditions,  confirming that models produce outputs that are syntactically compliant (when they succeed) but fail to anchor explanations in discriminative  visual evidence. Feature-based and Feature-value pairs conditions achieve the highest scores on most metrics, confirming that these intermediate formats are best handled by current MLLMs.

\begin{table*}[htbp!]
\centering
\caption{Mean LLM-as-a-judge semantic explainability scores (1--5\,$\uparrow$) %\cnat{cd usemos los scores en cada tabla poner flechita para arriba entre parentesis para indicar que higher is better - es convenio usual} 
by explanation condition, aggregated over 4~models, 4~datasets, and 6~few-shot
  configurations: $4\times26=104$ observations per cell.
  Values: mean~($\pm$SE). \textbf{Bold} = best per column.}
\label{tab:judge_cond_metric}
\small
\setlength{\tabcolsep}{4pt}
\renewcommand{\arraystretch}{1.1}
\begin{tabular}{@{}lccccccccc@{}}
\toprule
\textbf{Condition}
  & \textbf{TG} & \textbf{HF} & \textbf{CC} & \textbf{CP}
  & \textbf{Cn} & \textbf{S} & \textbf{LD} & \textbf{IF} & \textbf{LC} \\
\midrule
NLE
  & 3.62~{\tiny(.02)} & 4.46~{\tiny(.03)} & 4.68~{\tiny(.03)} & 4.95~{\tiny(.01)}
  & 4.81~{\tiny(.01)} & 3.73~{\tiny(.02)} & 3.69~{\tiny(.04)} & 4.70~{\tiny(.02)} & 4.84~{\tiny(.01)} \\
Features
  & 3.62~{\tiny(.03)} & \textbf{4.81}~{\tiny(.02)} & \textbf{4.68}~{\tiny(.03)} & \textbf{4.99}~{\tiny(.00)}
  & \textbf{4.97}~{\tiny(.01)} & 3.81~{\tiny(.03)} & 3.62~{\tiny(.04)} & \textbf{4.82}~{\tiny(.01)} & \textbf{4.89}~{\tiny(.01)} \\
Feature-value pairs
  & \textbf{3.70}~{\tiny(.03)} & 4.77~{\tiny(.02)} & 4.37~{\tiny(.04)} & 4.92~{\tiny(.01)}
  & 4.95~{\tiny(.01)} & \textbf{4.14}~{\tiny(.02)} & \textbf{3.91}~{\tiny(.04)} & 4.43~{\tiny(.03)} & 4.72~{\tiny(.02)} \\
DL Axioms
  & 2.31~{\tiny(.04)} & 4.40~{\tiny(.03)} & 4.20~{\tiny(.05)} & 3.97~{\tiny(.02)}
  & 4.94~{\tiny(.01)} & 2.85~{\tiny(.04)} & 3.10~{\tiny(.05)} & 3.05~{\tiny(.03)} & 2.97~{\tiny(.04)} \\
\bottomrule
\end{tabular}
\end{table*}
%\cnat{reemplazar todos los as-judge por as-a-judge - HECHO}

%Figure~\ref{fig:spearman_heatmap} shows the Spearman correlation between each judge metric and classification accuracy. Local Discriminativeness (LD) exhibits the strongest and most consistent correlation across conditions, demonstrating that models which successfully identify class-discriminative features are more likely to classify correctly. Logical Coherence (LC) is the strongest predictor for NLE, while most metrics show weak or near-zero correlation for DL Axioms, consistent with the degraded explanation quality observed for E5.

%\begin{figure}[htbp!]
%\centering
%\includegraphics[width=\linewidth]{figures/spearman_heatmap.png}
%\caption{Spearman correlation ($\rho$) between LLM-as-a-judge metrics and
  %classification accuracy across explanation conditions E2--E5.
  %Semantic explainability metric abbreviations in Table~\ref{tab:metrics}.}
%\label{fig:spearman_heatmap}
%\end{figure}

\begin{figure}[htbp!]
\centering
\includegraphics[width=0.85\linewidth]{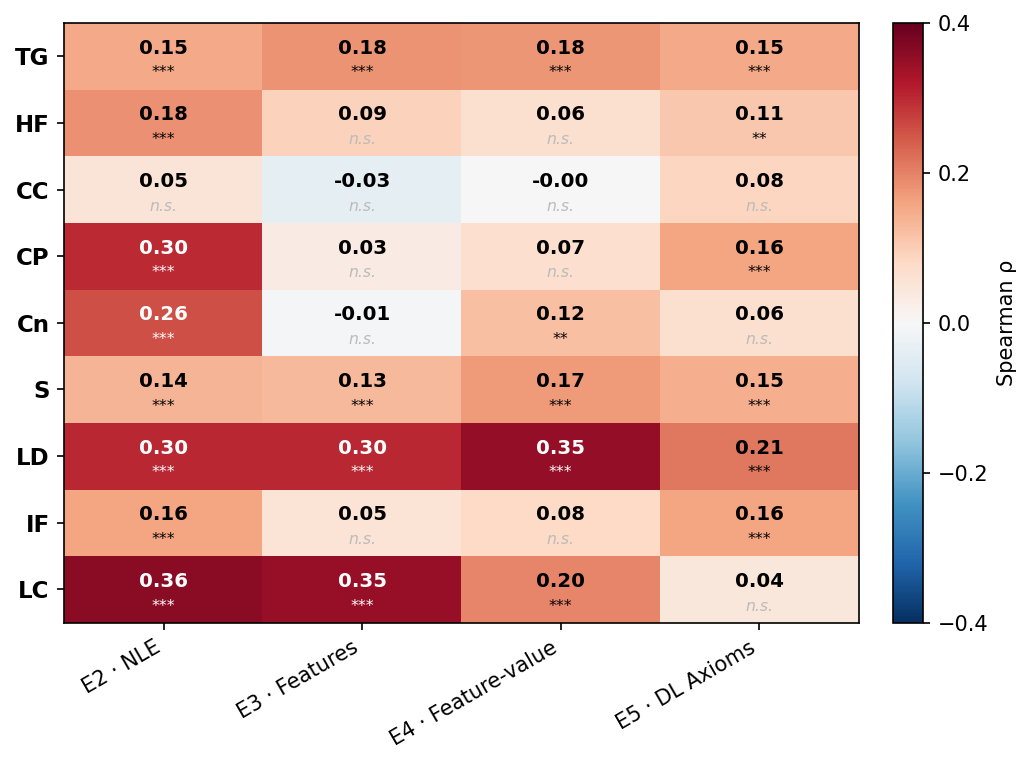}
\caption{Spearman $\rho$ between LLM-as-a-judge XAI metrics and classification
  accuracy across explanation conditions E2--E5.
  Significance markers indicate Bonferroni-corrected $p$-values over
  $9 \times 4 = 36$ tests: {*}~$p_{\mathrm{Bonf}}<0.05$,
  {**}~$p_{\mathrm{Bonf}}<0.01$,
  {***}~$p_{\mathrm{Bonf}}<0.001$;
  \textit{n.s.}~not significant.
  Metric abbreviations as in Table~\ref{tab:metrics}.}
\label{fig:spearman_heatmap}
\end{figure}

Figure~\ref{fig:spearman_heatmap} shows the Spearman correlation between
each judge metric and classification accuracy, with Bonferroni correction
over the 36 simultaneous tests ($9~\text{metrics} \times 4~\text{conditions}$).
Local Discriminativeness (LD) is the only metric that is both
consistently and strongly associated with accuracy across all four conditions
($\rho = 0.30$--$0.35$ for E2--E4, $\rho = 0.21$ for E5; all
$p_{\mathrm{Bonf}} < 10^{-10}$). Two further metrics---Textual Groundedness (TG)
and Specificity (S)---reach significance in every condition but with substantially
smaller magnitudes ($\rho \approx 0.13$--$0.18$), confirming LD as the
dominant predictor.
Logical Coherence (LC) is the strongest predictor for E2
($\rho = 0.36$, $p_{\mathrm{Bonf}} = 3 \times 10^{-35}$) and remains
significant for E3 and E4, but fails to reach significance for
E5 DL Axioms ($\rho = 0.04$, $p_{\mathrm{Bonf}} = 1.0$). More broadly,
under E5 three dimensions---Logical Coherence (LC), Concept Counting (CC), and
Conciseness (Cn)---are indistinguishable from zero after correction, and the
largest significant $\rho$ for E5 is only 0.21 (LD), versus 0.35--0.36 for
E2--E3. This pattern is consistent with the degraded explanation quality
observed for E5: when the explanation channel is unreliable, most of its
dimensions stop being diagnostic of the final classification decision.

% ============================================================
\section{Discussion}
\label{sec:discussion} 

\textbf{Explaining is Harder Than Predicting Alone.}
A central finding of our study is that forcing an MLLM to explain its decision does not improve classification accuracy; in fact, we observe the exact opposite. Accuracy drops monotonically as the complexity of the requested explanation increases, reaching its peak when no explanation is requested (E1). This suggests that deciding and explaining simultaneously is fundamentally more complex than deciding alone, contradicting the hypothesis that structured reasoning universally acts as a form of test-time regularisation.

\textbf{Accuracy does not imply explanation quality.}
A key observation from our preliminary experiments is that a model
can produce the correct class label while generating a factually
incorrect explanation---for instance, correctly identifying a flower
species while misreporting the number of petals visible in the image.
This decoupling motivates our two-track evaluation (accuracy +
judge scores) and underscores the importance of assessing explanation
quality independently of predictive performance.

\begin{figure}[htbp!]
\centering
\includegraphics[width=0.75\linewidth]{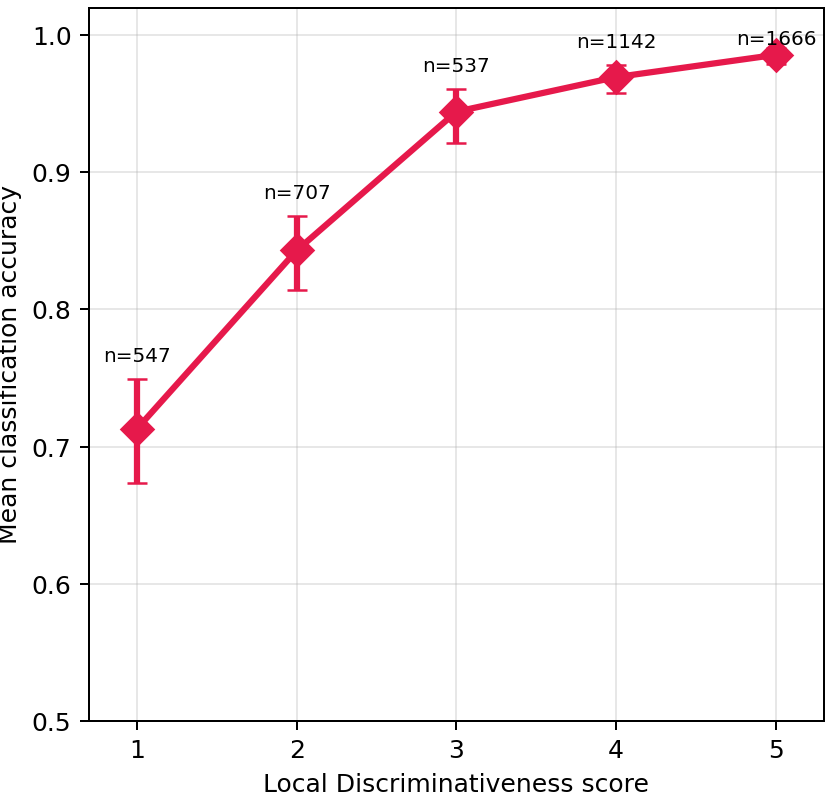}
\caption{Mean classification accuracy as a function of Local
  Discriminativeness (LD) score, pooled across all models, datasets,
  and explanation conditions (E2--E5).}
\label{fig:acc_vs_ld}
\end{figure}

\textbf{Discriminativeness as a proxy for correct decisions.}
Despite the drop in accuracy for highly concept-based explanations such as \dl Axioms, Local Discriminativeness (LD) emerges as the metric most strongly associated with correct predictions. When models succeed, they tend to justify their decisions using visual features that effectively separate the target class from alternatives, suggesting that access to discriminative evidence is closely tied to correctness. 
Model behaviour varies substantially across architectures and prompting conditions. For instance, Qwen3 VL 8B collapses on Textual Groundedness (TG) under E5 (TG\,=\,1.06 in Table~\ref{tab:judge_qwen}), whereas Gemini 2.5 Flash maintains comparatively stable performance across metrics and conditions, indicating that architectural and training differences significantly affect concept-based explanation capability beyond classification accuracy. An exception is LLaMA 4 Scout on DTD, where DL Axioms achieves the highest accuracy (86.1\%), surpassing the classification-only baseline (76.4\%), suggesting that structured axiom-style prompting may improve grounding for abstract texture classes.
Overall, these results admit a mechanistic interpretation. Across settings, improving access to discriminative information—e.g., by increasing the number of support images ($K{=}5$)—consistently enhances both accuracy and LD, as models become better at isolating features that distinguish the correct class. In contrast, increasing the number of candidate classes ($N$) systematically degrades both accuracy and LD, while leaving other explanation dimensions largely stable, indicating that the effect is specific to discriminative reasoning. 
Together, this suggests that LD is a strong proxy for whether the model has correctly identified class-distinguishing evidence (Figure~\ref{fig:acc_vs_ld}): when such evidence is successfully extracted, the model not only predicts correctly but can also verbalise the underlying rationale.

%\textbf{Discriminativeness as a proxy for correct decisions.}
%Despite the drop in accuracy for highly concept-based explanations like \dl Axioms, Local Discriminativeness (LD) emerged as the metric that most strongly correlates with correct predictions. When a model successfully justifies its decision using visual features that effectively separate the chosen class from the alternatives, it is more likely to be correct. This implies that detecting these discriminative features is a prerequisite for a correct decision; once detected, the model is capable of verbalising them.
%Notably, explanation quality varies substantially across models: Qwen3 VL 8B collapses on Textual Groundedness under E5 (TG\,=\,1.06 in Table~\ref{tab:judge_qwen}), while Gemini 2.5 Flash maintains comparatively robust scores across all conditions, suggesting that architectural and training differences substantially impact concept-based explanation capability beyond raw classification performance.
%An exception is LLaMA 4 Scout on DTD, where DL Axioms achieves the highest accuracy (86.1\%), surpassing the classification-only baseline (76.4\%). We hypothesise that the formal axiom structure forces more precise grounding of abstract textural features, partially compensating for the model's weaker baseline on this dataset. 

\textbf{The Instruction-Tuning Hypothesis.}
We hypothesise that the capability gap between predicting and formally explaining stems from how these models are instruction-tuned. While current MLLMs excel at concrete tasks like classification, they are not explicitly optimised to generate formal logical representations in tandem with classification. Asking a model to perform both tasks simultaneously within a single prompt overcomplicates generation, a phenomenon aligned with recent work showing that forcing explicit reasoning steps can sometimes degrade performance by biasing models toward generalised rules rather than contextual evidence \cite{liu2025mindyourstep}. %Interestingly, feature-based (E3) and feature–value pair (E4) conditions achieve similar scores across most metrics, despite E4 imposing substantially more structural constraints. This suggests that increased formalisation does not translate into performance gains and may be introducing additional complexity that models do not handle effectively.
 This is further evidenced by the contrasting pattern observed under E5: while Hallucination Free (HF) and Conciseness (Cn) scores remain high (4.40 and 4.94 respectively), Textual Groundedness (TG), Local Discriminativeness (LD), and Logical Coherence (LC) collapse — suggesting that models learn to reproduce the \emph{syntactic structure} of DL reasoning without filling it with discriminative visual content. This mirrors a broader question central to compositional AI: whether models truly compose symbolic representations or merely imitate their surface form~\cite{huang2025mimicking}.

\textbf{Limitations.} Our evaluation framework has several constraints. First, our results are limited to four MLLMs and four datasets. Additionally, due to the independent nature of episodic \icl trials, models do not accumulate knowledge iteratively across queries. This setup makes it fundamentally difficult to assess true ``zero knowledge'' prior to prompting, or to perform ablation studies that guarantee a complete lack of data leakage from the models' pre-training phase. Consequently, separating genuine in-context reasoning from pre-training memory retrieval remains an inherent limitation of this paradigm --- further compounded by the fact that LLMs cannot fully suppress pre-trained knowledge when processing in-context demonstrations, with smaller models tending to rely more heavily on label semantics than on in-context mappings~\cite{zhang2024mystery}, and overcoming strong prior biases remaining an open challenge~\cite{dong2024survey}. To mitigate this, our prompts explicitly instruct the model not to rely on external world knowledge or speculative attributes (see Appendix~\ref{app:prompts}), grounding classification and explanations in the visual evidence of the support images. A further limitation concerns classes whose discriminative features cannot be described in absolute terms --- that is, features that only become discriminative \emph{relative} to the other classes present in the support set. For instance, subtle breed-specific coat markings in Oxford-IIIT Pets may only be identifiable as class-defining when compared against the other breeds in the trial, not in isolation. In such cases, Local Discriminativeness (LD) --- which is
evaluated by the judge without access to the support set ---
could in principle underestimate explanation quality for
classes whose discriminative features are only articulable
relative to the specific episode context.
However, our ablation study (Appendix~\ref{app:judge_validation})
shows empirically that LD rankings are robust to support-image
access ($\rho = 0.627$, mean score shift $= -0.049$ points),
suggesting that the judge's prior knowledge of the candidate
classes is sufficient for reliable LD assessment within the
benchmarks used in this study.
Addressing contrastive-dependent features more generally would
require \emph{Global Discriminativeness} (GD), which evaluates
whether explanations highlight features that are truly
pathognomonic at the dataset level
(see Appendix~\ref{app:metrics}); however, GD requires
knowledge beyond a single trial and is therefore left for
future work.

% ============================================================
\section{Conclusion and Future Work}
\label{sec:conclusion}

In this paper, we presented a systematic empirical study evaluating whether frozen MLLMs can produce structured, semantically rich, and verifiable explanations for their in-context image classification decisions. 
Our findings demonstrate that explaining is genuinely harder than predicting alone. Contrary to the hypothesis that forcing explicit reasoning acts as a test-time regularisation that improves accuracy, we observe a monotonic performance drop: classification accuracy degrades from 93.8\% to 90.1\% as the required explanations become more formally rigorous (from simple labels to Description Logic ontological axioms). 

Furthermore, our LLM-as-a-judge evaluation reveals that classification accuracy does not imply explanation quality. While current models struggle significantly to articulate formal logic structures, their ability to identify and verbalise class-discriminative visual features (Local Discriminativeness) is the strongest predictor of a correct classification. 

Future work should explore methods to evaluate contrastive-dependent visual features, exploit the machine-verifiable nature of DL axioms --- generated under E5 --- as structured inputs for a secondary safe classifier, and investigate whether instruction-tuning can be explicitly optimised for formal explainability without compromising predictive performance.

%\cnat{al terminar verificar qeu todas las siglas se explican la 1a vez q se usan, por ej, faltan SFT y RLHF -- se agregaron ver el resto}
Ultimately, we argue that just as \textit{current models} are optimized via Supervised Fine-Tuning (SFT) and Reinforcement Learning from Human Feedback (RLHF) to align with human intent, future frameworks\textit{ should integrate concept-based explanations} directly into the optimization objective. This approach creates a crucial bidirectional link between semantics—the 'why' for human understanding—and formality—the mathematical 'proof' for system integrity. By leveraging these explanations to reach a formal state of explainability, we move beyond simple task performance toward a new generation of models that are inherently and provably transparent.

Code, prompts, and evaluation pipeline are publicly available at \url{https://github.com/letyrodridc/research-llms-in-context-learning-explained}.

% ============================================================
\textbf{Impact Statement} This paper advances explainable AI and ICL by contributing tools and evaluation frameworks for auditing AI decision-making. We foresee no immediate negative societal consequences beyond those generally associated with advancing large language model capabilities.

\section*{Acknowledgements} This work was supported by the TSI-100927-2023-1 project under the Recovery, Transformation and Resilience Plan funded by the European Union NextGenerationEU through the Ministry for Digital Transformation and the Civil Service, and grants PID2023-149128NBI00 and PID2023-150070NB-I00 funded by MICIU/AEI /10.13039/501100011033 and by ERDF, EU.

%  'SIGNNET: TOWARDS DEMOCRATIZING CONTENT ACCESSIBILITY FOR THE DEAF BY ALIGNING MULTI-MODAL SIGN REPRESENTATIONS'
% \section*{Acknowledgements}
% Upon acceptance.

\bibliography{references}
\bibliographystyle{icml2026}

% ============================================================
\newpage
\appendix

\section{Prompt Templates}
\label{app:prompts}

This appendix provides the exact prompt templates used in our experiments. Variables populated at runtime are enclosed in curly braces, such as \texttt{\{CONDITION\_INSTRUCTION\}}.

\subsection{Classification Prompts}
All MLLM classifiers were evaluated using a shared system prompt, into which specific explanation formatting constraints were injected based on the evaluated condition.

\vspace{1em}
\noindent\textbf{SHARED\_SYSTEM\_PROMPT}
\vspace{0.3em}\hrule\vspace{0.5em}
{\small
\textbf{\# TASK}
\begin{enumerate}[leftmargin=*, nosep]
    \item Analyze the provided labeled examples to identify the observable visual features that distinguish each class.
    \item Examine the Query Image and compare it strictly against the provided examples.
    \item Determine which label from the provided options best matches the Query Image.
\end{enumerate}

\vspace{0.5em}
\textbf{\# INSTRUCTIONS}
\begin{itemize}[leftmargin=*, nosep]
    \item Base your decision only on observable visual evidence in the examples and the Query Image.
    \item Use the examples to infer discriminative visual patterns for each class.
    \item Do not use external world knowledge, hidden assumptions, or speculative attributes.
    \item Use only labels from the provided final options list.
    \item Choose exactly one final label.
    \item Output only the requested XML tags.
    \item Do not output any text outside the XML tags.
    \item Do not output any additional XML tags other than those explicitly defined in \texttt{\{CONDITION\_INSTRUCTION\}}.
\end{itemize}

\vspace{0.5em}
\texttt{\{CONDITION\_INSTRUCTION\}}

\vspace{0.5em}
\texttt{<response>}final\_class\texttt{</response>}

\vspace{0.5em}
The content of \texttt{<response>} must be exactly one label copied verbatim from the provided options list.
}
\vspace{0.5em}\hrule\vspace{1em}

The \texttt{\{CONDITION\_INSTRUCTION\}} variable varies according to the explanation setting as follows:

\vspace{1em}
\noindent\textbf{CONDITION INSTRUCTIONS BY EXPLANATION TYPE}
\vspace{0.3em}\hrule\vspace{0.5em}
{\small
\textbf{E1 ---Classification with no explanation (Baseline):} Provide only the final class label inside the response tag.

\vspace{0.5em}
\textbf{E2 --- NLE:} Write a concise natural language explanation grounded only in observable visual evidence.\\
\texttt{<explanation>}\\
Brief explanation of why the visual evidence supports the selected class.\\
\texttt{</explanation>}

\vspace{0.5em}
\textbf{E3 --- Features:} Extract only the smallest sufficient set of critical, concrete, observable visual features needed to support the classification.\\
- Use short noun phrases only.\\
- Do not use abstract interpretations, causal explanations, narrative reasoning, or full sentences inside the features tag.\\
- Use bullet points:\\
\texttt{<features>}\\
- short observable feature\\
- short observable feature\\
\texttt{</features>}

\vspace{0.5em}
\textbf{E4 --- Feature-value pairs:} First list the critical observable features present only in the Query Image. Then formulate a knowledge base (KB) with a small set of IF-THEN rules grounded in observable patterns from the labeled examples in the support images and the Query Image, for all classes. Try to unify the conditions so that there is only one rule per class. Rules must refer only to visible features. Then check which rule(s) are best satisfied by the Query Image.\\
\texttt{<features>}\\
- observable feature: feature value\\
\texttt{</features>}\\
\texttt{<kb>}\\
- IF [observable visual features = feature value AND/OR observable visual features = feature value] THEN [class1]\\
- IF [observable visual features = feature value] THEN [class2]\\
\texttt{</kb>}\\
\texttt{<rule\_check>}\\
- State which rule(s) from the kb fires for the Query Image and briefly explain why, referring only to the features listed above. Do not discuss other rules or introduce new features.\\
\texttt{</rule\_check>}

\vspace{0.5em}
\textbf{E5 --- DL Axioms:} Represent the classification evidence using minimal and concise Description Logic statements.\\
\textbf{\# WHAT TO WRITE}\\
- In \texttt{<tbox>}, define the formal class conditions (Necessary, Sufficient, or Necessary \& Sufficient) using Description Logic (DL) axioms in Web Ontology Language (OWL 2) derived from the labelled examples to describe which visual features are associated with each class.\\
- In \texttt{<abox>}, state only the specific property assertions in DL OWL 2 observed in the Query Image to record which visual features are present.\\
- In \texttt{<dl\_explanation>}, list only the exact TBox rules triggered by the ABox assertions to classify the Query Image. Do not use natural language---include only the fired TBox rules.

\vspace{0.5em}
\textbf{\# HOW TO INTERPRET THE STATEMENTS}\\
- \texttt{[Class1] $\sqsubseteq$ hasVisualFeature([F1])} means: images of this class tend to have this visual feature in the provided examples (necessary condition).\\
- \texttt{hasVisualFeature([F1]) $\sqsubseteq$ [Class1]} means: observing this visual feature is enough to classify the image as from this class in the provided examples (Sufficient condition).\\
- \texttt{[Class1] $\equiv$ hasVisualFeature([F1]) $\sqcap$ hasVisualFeature([F2])} means: this combination of features defines this class in the provided examples (Necessary and sufficient condition).

\vspace{0.5em}
\textbf{\# RULES}\\
- Use only observable visual features. Do not use hidden properties, external knowledge, or speculation. Treat all axioms as visually grounded in the provided examples. Keep the output concise.\\
\texttt{<tbox>}\\
- \texttt{[Class1] $\sqsubseteq$ hasVisualFeature([F1])}\\
- \texttt{hasVisualFeature([F2]) $\sqsubseteq$ [Class2]}\\
- \texttt{[Class3] $\equiv$ hasVisualFeature([F3]) $\sqcap$ hasVisualFeature([F4])}\\
\texttt{</tbox>}\\
\texttt{<abox>}\\
- \texttt{hasVisualFeature(Query, [F3])}\\
- \texttt{hasVisualFeature(Query, [F4])}\\
\texttt{</abox>}\\
\texttt{<dl\_explanation>}\\
- Rule(s) fired: \texttt{[Class3] $\equiv$ hasVisualFeature([F3]) $\sqcap$ hasVisualFeature([F4])}\\
\texttt{</dl\_explanation>}
}
\vspace{0.5em}\hrule\vspace{1em}

\subsection{LLM-as-a-judge Prompts}

\vspace{1em}
\noindent\textbf{JUDGE\_PROMPT}
\vspace{0.3em}\hrule\vspace{0.5em}
{\small
You are a fair LLM judge. Your task is providing clear, objective feedback based on specific criteria to evaluate the quality of an explanation generated by an independent LLM query regarding a previous image classification decision. Please, ensure that each assessment reflects the absolute dimensions to score.

You will receive:
1. The Query image.
2. The set of candidate class labels.
3. The candidate model output (explanation).
4. The predicted class.
5. A description of the explanation condition that specifies the type of explanation the model was asked to produce and how its output should be interpreted.

To give feedback, rate the explanation on each of all given dimensions writing a score that is an integer between 1 and 5.

\vspace{0.5em}
\textbf{\# IMPORTANT EVALUATION PRINCIPLES}\\
- Evaluate the explanation itself, not whether the response is correct or incorrect.\\
- Evaluate whether the explanation supports the predicted class, regardless of whether the prediction is ultimately correct with respect to external ground truth.\\
- Judge only what is explicitly present in the model output.\\
- Use the Query image as the source of truth for whether mentioned features are visible.

\vspace{0.5em}
\textbf{\# CONDITION DESCRIPTION}\\
\texttt{\{CONDITION\_DESCRIPTION\}}

\vspace{0.5em}
\textbf{\# DIMENSIONS TO SCORE}\\
\textit{(1. Textual Groundedness, 2. Hallucination free, 3. Concept counting, 4. Comprehensibility, 5. Conciseness, 6. Specificity, 7. Discriminativeness, 8. Instruction following, 9. Logical coherence. The complete 1--5 scoring rubrics for each metric provided to the judge are detailed in Appendix~\ref{app:metrics}.)}

\vspace{0.5em}
\textbf{\# OUTPUT INSTRUCTIONS}\\
Return your evaluation using exactly the following XML structure:\\
\texttt{<evaluation>}\\
\texttt{~~<textual\_groundedness>1-5</textual\_groundedness>}\\
\texttt{~~...}\\
\texttt{~~<logical\_coherence>1-5</logical\_coherence>}\\
\texttt{</evaluation>}\\
Please, do not generate any other opening, closing, and explanations.
}
\vspace{0.5em}\hrule\vspace{1em}

\noindent\textbf{JUDGE CONDITION DESCRIPTIONS}
\vspace{0.3em}\hrule\vspace{0.5em}
{\small
The \texttt{\{CONDITION\_DESCRIPTION\}} variable adapts to the evaluated explanation type:

\textbf{E2 --- NLE:} The model was asked to classify the Query image and provide a concise natural-language explanation. Expected format: Only \texttt{<explanation>} and \texttt{<response>} tags. The explanation should be grounded only in observable visual evidence and written in free-form natural language.

\textbf{E3 --- Features:} The model was asked to classify the Query image and provide a minimal list of critical observable features supporting the classification. Expected format: Only \texttt{<features>} and \texttt{<response>} tags. Features should be short noun phrases formatted as bullet points, without abstract interpretations or full sentences.

\textbf{E4 --- Feature-value pairs:} The model was asked to classify the Query image and provide a rule-based explanation. Expected format: \texttt{<features>}, \texttt{<kb>}, \texttt{<rule\_check>}, and \texttt{<response>} tags. Rules in \texttt{<kb>} must be IF-THEN grounded in observable patterns covering all classes. \texttt{<rule\_check>} should state which rule fires for the Query image referring only to the listed features.

\textbf{E5 --- DL Axioms:} The model was asked to classify the Query image and provide a Description Logic (DL)-based explanation using OWL 2 axioms. Expected format: \texttt{<tbox>}, \texttt{<abox>}, \texttt{<dl\_explanation>}, and \texttt{<response>} tags. The \texttt{<tbox>} defines formal class conditions ($\sqsubseteq, \equiv, \sqcap$). The \texttt{<abox>} lists specific property assertions in the Query image. The \texttt{<dl\_explanation>} lists exactly the TBox rules fired by the ABox assertions without natural language.
}
\vspace{0.5em}\hrule\vspace{1em}

\section{Models Details}
\label{app:models}
All models are accessed via the OpenRouter API in inference-only mode,
with temperature set to 0 to ensure deterministic outputs and no
fine-tuning or weight updates applied.

\textbf{Gemini 2.5 Flash}~\citep{gemini25flash2025} is a
state-of-the-art proprietary model from Google DeepMind with
advanced multimodal reasoning capabilities and a 1M-token context
window. It serves as the high-performance proprietary baseline.

\textbf{Gemma 4 26B}~\citep{gemma4_2026} is an open-weight
Mixture-of-Experts model from Google DeepMind (26B total parameters,
4B active per forward pass) with a 262K-token context window,
included for its competitiveness on multimodal
tasks~\citep{ramachandran2025evaluating}.

\textbf{Qwen3 VL 8B}~\citep{qwen3vl2025} is an open-weight dense
model from Alibaba Cloud (8.77B parameters) with early-fusion visual
tokenisation and a 131K-token context window.
It represents the most parameter-efficient model in our selection.

\textbf{LLaMA 4 Scout}~\citep{meta2025llama4} is an open-weight
Mixture-of-Experts model from Meta (109B total parameters, 17B
active per forward pass) with native multimodal support and a
328K-token context window via OpenRouter~\citep{meta2025llama4}.
It is widely used as a baseline in recent multimodal studies.

\section{Semantic Explainability Metric Definitions}
\label{app:metrics}

%\cnat{veo formato demasiado espaciadoen las metricas, como tabulaciones de mas?}

% --- MÉTRICA 1 ---
\subsection*{1. Textual Groundedness}
\textbf{Desideratum:} Every relevant concept in the image is mentioned in the text explanation.

\textbf{Definition (IMG $\rightarrow$ NLE):} A well grounded vision-to-text explanation mentions every relevant concept in the image in the explanation.

\textbf{What it measures:} The degree to which the generated text captures all salient objects, attributes, and spatial relationships present in the source image. Whether the text provides an exhaustive account of the visual scene without omitting relevant concepts (\cite{yang2022improving}).

\textbf{Scores:}
\begin{itemize}
    \item \textbf{Not Grounded:} Severe omissions of key concepts in the explanation.
    \item \textbf{Weakly Grounded:} Major omissions of key concepts.
    \item \textbf{Moderately Grounded:} Few omissions of key concepts.
    \item \textbf{Strongly Grounded:} One omission of key concepts.
    \item \textbf{Perfectly Grounded:} Zero omissions of key concepts.
\end{itemize}

% --- MÉTRICA 2 ---
\subsection*{2. Hallucination-Free}
\textbf{Desideratum:} Every relevant concept in the explanation is visible in the image.

\textbf{Definition (NLE $\rightarrow$ IMG):} A hallucination-free explanation reflects accurate visual evidence.

\textbf{What it measures:} Evaluates whether the explanation reflects accurate visual evidence (any claims made in the explanation must be visually verifiable in the query image without hallucinating non-existent features).

\textbf{Scores:}
\begin{itemize}
    \item \textbf{Severe hallucinations}
    \item \textbf{Major hallucinations}
    \item \textbf{Few hallucinations}
    \item \textbf{One hallucination}
    \item \textbf{Hallucination-free:} Zero hallucinations.
\end{itemize}

% --- MÉTRICA 3 ---
\subsection*{3. Concept Counting}
\textbf{Desideratum:} The explanation accurately quantifies the exact number of relevant concepts present in the image.

\textbf{Definition:} A precisely counted explanation ensures that the elements counted in the NLE faithfully correspond with the number of concepts in the image.

\textbf{What it measures:} Evaluates the quantitative accuracy of the explanation (penalizing any discrepancies between the stated count of elements in the text and the actual number of countable elements visible in the image).

\textbf{\textit{Note}:} This metric is different from hallucination-free because it does reflect, for instance, a model explaining a flower of class $f$ has 5 petals while the query image shows 6 petals.

\textbf{Scores:}
\begin{itemize}
    \item \textbf{Wrong concept counting:} 0\% of the countable features are well counted.
    \item \textbf{Weak concept counting:} 1–33\% of the countable features are well counted.
    \item \textbf{Moderate concept counting:} 34–66\% of the countable features are well counted.
    \item \textbf{Almost perfect concept counting:} 67–99\% of the countable features are well counted.
    \item \textbf{Perfect concept counting:} 100\% of the countable features are well counted.
\end{itemize}

% --- MÉTRICA 4 ---
\subsection*{4. Comprehensibility}
\textbf{Desideratum:} The explanation is simple, readable, and perfectly accessible to an end-user without unnecessary complexity.

\textbf{Definition:} A highly comprehensible explanation is easy to read, simple, and accessible to end-users.

\textbf{What it measures:} Evaluates whether: 1) explanations are simple and accessible to end-users by measuring readability and penalizing overly complex or unnecessarily sophisticated explanations; 2) descriptions in terms of symbols are semantically and structurally similar to those of human experts \cite{arrieta2020explainable}.

\textbf{Scores:}
\begin{itemize}
    \item \textbf{Not Comprehensible:} Highly complex.
    \item \textbf{Weak Comprehensibility:} Mostly complex.
    \item \textbf{Medium Comprehensibility:} Somewhat complex.
    \item \textbf{Strong Comprehensibility:} A light degree of complexity.
    \item \textbf{Perfectly Comprehensible:} No complexity.
\end{itemize}

% --- MÉTRICA 5 ---
\subsection*{5. Conciseness}
\textbf{Desideratum:} The explanation conveys only the strictly necessary information efficiently.

\textbf{Definition:} A concise explanation conveys the necessary information efficiently without unnecessary length nor fillers.

\textbf{What it measures:} Evaluates the length of the explanation (penalizing unnecessarily verbose answers, filler words, or repetition).

\textbf{Scores:}
\begin{itemize}
    \item \textbf{Not Concise:} Highly verbose, most content is redundant or removable.
    \item \textbf{Weakly Concise:} Mostly verbose, several sentences can be removed without loss of meaning.
    \item \textbf{Medium Conciseness:} Somewhat verbose, few sentences or parts can be removed while preserving full meaning.
    \item \textbf{Strongly Concise:} Light verbosity, one sentence can be removed without losing meaning.
    \item \textbf{Perfectly Concise:} No unnecessary verbosity, nothing can be removed without losing meaning.
\end{itemize}

% --- MÉTRICA 6 ---
\subsection*{6. Specificity}
\textbf{Desideratum:} The explanation relies exclusively on precise, exact, and non-generic details about the sample being classified.

\textbf{Definition:} A highly specific explanation provides precise, concrete, and non-generic information about the classified subject. This metric is evaluated locally, meaning it only considers the individual sample at hand, without requiring knowledge of other classes.

\textbf{What it measures:} Evaluates whether the explanation relies on precise details on a local sample-basis (penalizing vague, broad, or generic statements).

\textbf{Scores:}
\begin{itemize}
    \item \textbf{Not Specific:} Highly vague, 0\% of all features are precise, concrete, and non-generic.
    \item \textbf{Weakly Specific:} Mostly vague, 1–33\% of all features are precise, concrete, and non-generic.
    \item \textbf{Moderately Specific:} Somewhat vague, 34–66\% of all features are precise, concrete, and non-generic features.
    \item \textbf{Strongly Specific:} Minimal vagueness, 67–99\% of all features are precise, concrete, and non-generic.
    \item \textbf{Perfectly Specific:} Zero vagueness, 100\% of all features are precise, concrete, and non-generic.
\end{itemize}

% --- MÉTRICA 7 ---
\subsection*{7. Global Discriminativeness}
%\cnat{en lugar de poner future work (era para nosotrso) incluir una frase al final del appendix diciendo que dejamos metrica x y y para futuros experimentos considerando clases beyond those seen in a single independent trial in ICL.}
\textbf{Desideratum:} The explanation successfully isolates the unique features that separate the predicted class from all others.

\textbf{Definition:} A discriminative explanation highlights the features and values that truly support the predicted class over all others — focusing on pathognomonic features rather than correlated or commonly shared ones. This metric is evaluated globally: the evaluator must be aware of all classes present in the dataset.

\textbf{What it measures:} Evaluates the capability of providing global key reasons causing a prediction.

\textbf{Scores:}
\begin{itemize}
    \item \textbf{Not Discriminative:} 0\% pathognomonic features.
    \item \textbf{Weakly Discriminative:} 1–33\% pathognomonic features.
    \item \textbf{Moderately Discriminative:} 34–66\% pathognomonic features.
    \item \textbf{Strongly Discriminative:} 67–99\% pathognomonic features.
    \item \textbf{Perfectly Discriminative:} 100\% pathognomonic features.
\end{itemize}

% --- MÉTRICA 8 ---
\subsection*{8. Local Discriminativeness}
\textbf{Desideratum:} The explanation successfully isolates the unique features that separate the predicted class from all others.

\textbf{Definition:} A discriminative explanation highlights the features and values that truly support the predicted class over all others, evaluated locally within the trial.

\textbf{What it measures:} Evaluates the capability of providing local key reasons causing a prediction.

\textbf{Scores:}
\begin{itemize}
    \item \textbf{Not Discriminative:} 0\% pathognomonic features.
    \item \textbf{Weakly Discriminative:} 1–33\% pathognomonic features.
    \item \textbf{Moderately Discriminative:} 34–66\% pathognomonic features.
    \item \textbf{Strongly Discriminative:} 67–99\% pathognomonic features.
    \item \textbf{Perfectly Discriminative:} 100\% pathognomonic features.
\end{itemize}

% --- MÉTRICA 9 ---
\subsection*{9. Instruction Following}
\textbf{Desideratum:} The output flawlessly satisfies all explicit structure and format constraints.

\textbf{Definition:} An explanation that follows instructions is compliant with open-ended and format instructions \cite{zeng2023evaluating}.

\textbf{What it measures:} Evaluates how closely the model's output follows all explicit instructions and constraints in the prompt.

\textbf{Scores:}
\begin{itemize}
    \item \textbf{No Adherence:} 0\% followed.
    \item \textbf{Weak Adherence:} 1–33\% followed.
    \item \textbf{Partial Adherence:} 34–66\% followed.
    \item \textbf{Strong Adherence:} 67–99\% followed.
    \item \textbf{Perfect Adherence:} 100\% followed.
\end{itemize}

% --- MÉTRICA 10 ---
\subsection*{10. Logical Coherence}
\textbf{Desideratum:} The sentences connect logically to form a valid, smooth, and complete deduction.

\textbf{Definition:} A logically coherent explanation builds from sentence to sentence into a valid body of information \cite{liu2023g}.

\textbf{What it measures:} Evaluates whether the explanation forms a valid logical progression.

\textbf{Scores:}
\begin{itemize}
    \item \textbf{Not Coherent:} Completely disconnected clauses.
    \item \textbf{Weakly Coherent:} Severe logical errors.
    \item \textbf{Moderately Coherent:} Some logical issues.
    \item \textbf{Strongly Coherent:} Mostly smooth, minor error.
    \item \textbf{Perfectly Coherent:} Flawless deduction.
\end{itemize}

% --- MÉTRICA 11 ---
\subsection*{11. Explanation Reproducibility}
\textbf{Desideratum:} The core reasoning remains consistent across identical runs.

\textbf{Definition:} A reproducible explanation maintains consistency in extracted features across repeated runs.

\textbf{What it measures:} Evaluates stability across runs via feature overlap.

\textbf{Scores:}
\begin{itemize}
    \item \textbf{Not Reproducible:} 0\% overlap.
    \item \textbf{Weakly Reproducible:} 1–33\% overlap.
    \item \textbf{Moderately Reproducible:} 34–66\% overlap.
    \item \textbf{Strongly Reproducible:} 67–99\% overlap.
    \item \textbf{Perfectly Reproducible:} 100\% overlap.
\end{itemize}

\paragraph{Note on evaluation scope.}
Global discriminativeness and explanation reproducibility are included for completeness but are not evaluated in the present study. Both require information beyond a single independent trial in the in-context learning setting (e.g., access to the full class set or repeated runs), and are therefore left for future work.

\section{Detailed Results per Model}
\label{app:extra}

Tables~\ref{tab:full_gemini}--\ref{tab:full_llama} report the full
accuracy breakdown per model across all datasets, $N$-way settings,
and $K$-shot settings. Observations per cell vary by $N$: $N{=}2 \to 6$ observations, $N{=}3 \to 4$ observations, $N{=}4 \to 3$ observations, $Q{=}1$.
$^\dagger$Classification-only baseline.

% ---- Table A1: Gemini 2.5 Flash ----
\begin{table*}[htbp!]
\centering
\caption{Full FS-ICL classification accuracy breakdown for model Gemini 2.5 Flash (\%). $N\in\{2,3,4\}$, $K\in\{1,5\}$, $Q{=}1$.
  $^\dagger$Classification-only baseline.}
\label{tab:full_gemini}
\scriptsize
\setlength{\tabcolsep}{2.5pt}
\renewcommand{\arraystretch}{1.1}
\begin{tabular}{@{}l
    cc cc cc
    cc cc cc
    cc cc cc
    cc cc cc@{}}
\toprule
& \multicolumn{6}{c}{\textbf{Flowers}}
& \multicolumn{6}{c}{\textbf{Oxford Pets}}
& \multicolumn{6}{c}{\textbf{CIFAR-10}}
& \multicolumn{6}{c}{\textbf{DTD}} \\
\cmidrule(lr){2-7}\cmidrule(lr){8-13}\cmidrule(lr){14-19}\cmidrule(lr){20-25}
& \multicolumn{2}{c}{2-way}
& \multicolumn{2}{c}{3-way}
& \multicolumn{2}{c}{4-way}
& \multicolumn{2}{c}{2-way}
& \multicolumn{2}{c}{3-way}
& \multicolumn{2}{c}{4-way}
& \multicolumn{2}{c}{2-way}
& \multicolumn{2}{c}{3-way}
& \multicolumn{2}{c}{4-way}
& \multicolumn{2}{c}{2-way}
& \multicolumn{2}{c}{3-way}
& \multicolumn{2}{c}{4-way} \\
\cmidrule(lr){2-3}\cmidrule(lr){4-5}\cmidrule(lr){6-7}
\cmidrule(lr){8-9}\cmidrule(lr){10-11}\cmidrule(lr){12-13}
\cmidrule(lr){14-15}\cmidrule(lr){16-17}\cmidrule(lr){18-19}
\cmidrule(lr){20-21}\cmidrule(lr){22-23}\cmidrule(lr){24-25}
\textbf{Explan. Condition}
  & 1s & 5s & 1s & 5s & 1s & 5s
  & 1s & 5s & 1s & 5s & 1s & 5s
  & 1s & 5s & 1s & 5s & 1s & 5s
  & 1s & 5s & 1s & 5s & 1s & 5s \\
\midrule
Classif.$^\dagger$
  & 100.0 & 100.0 & 100.0 & 100.0 & 100.0 & 100.0
  & 100.0 & 100.0 & 100.0 & 100.0 & 91.7 & 100.0
  & 100.0 & 100.0 & 91.7 & 100.0 & 91.7 & 91.7
  & 91.7 & 100.0 & 100.0 & 100.0 & 75.0 & 91.7 \\
NLE
  & 100.0 & 100.0 & 100.0 & 100.0 & 100.0 & 100.0
  & 100.0 & 100.0 & 100.0 & 100.0 & 100.0 & 100.0
  & 100.0 & 100.0 & 91.7 & 100.0 & 91.7 & 91.7
  & 83.3 & 100.0 & 100.0 & 100.0 & 75.0 & 100.0 \\
Features
  & 100.0 & 100.0 & 100.0 & 100.0 & 100.0 & 100.0
  & 100.0 & 100.0 & 100.0 & 100.0 & 100.0 & 100.0
  & 100.0 & 100.0 & 91.7 & 100.0 & 91.7 & 91.7
  & 75.0 & 100.0 & 100.0 & 100.0 & 75.0 & 100.0 \\
Feature-value pairs
  & 100.0 & 100.0 & 100.0 & 100.0 & 100.0 & 100.0
  & 100.0 & 100.0 & 100.0 & 100.0 & 83.3 & 100.0
  & 100.0 & 100.0 & 91.7 & 100.0 & 91.7 & 91.7
  & 83.3 & 100.0 & 91.7 & 100.0 & 75.0 & 91.7 \\
DL Axioms
  & 100.0 & 100.0 & 100.0 & 100.0 & 100.0 & 100.0
  & 100.0 & 100.0 & 100.0 & 100.0 & 91.7 & 100.0
  & 100.0 & 100.0 & 91.7 & 100.0 & 91.7 & 91.7
  & 83.3 & 100.0 & 91.7 & 100.0 & 83.3 & 83.3 \\
\bottomrule
\end{tabular}
\end{table*}

% ---- Table A2: Gemma 4 26B ----
\begin{table*}[htbp!]
\centering
\caption{Full FS-ICL classification accuracy breakdown for model Gemma 4 26B (\%). $N\in\{2,3,4\}$, $K\in\{1,5\}$, $Q{=}1$.
  $^\dagger$Classification-only baseline.}
\label{tab:full_gemma}
\scriptsize
\setlength{\tabcolsep}{2.5pt}
\renewcommand{\arraystretch}{1.1}
\begin{tabular}{@{}l
    cc cc cc
    cc cc cc
    cc cc cc
    cc cc cc@{}}
\toprule
& \multicolumn{6}{c}{\textbf{Flowers}}
& \multicolumn{6}{c}{\textbf{Oxford Pets}}
& \multicolumn{6}{c}{\textbf{CIFAR-10}}
& \multicolumn{6}{c}{\textbf{DTD}} \\
\cmidrule(lr){2-7}\cmidrule(lr){8-13}\cmidrule(lr){14-19}\cmidrule(lr){20-25}
& \multicolumn{2}{c}{2-way}
& \multicolumn{2}{c}{3-way}
& \multicolumn{2}{c}{4-way}
& \multicolumn{2}{c}{2-way}
& \multicolumn{2}{c}{3-way}
& \multicolumn{2}{c}{4-way}
& \multicolumn{2}{c}{2-way}
& \multicolumn{2}{c}{3-way}
& \multicolumn{2}{c}{4-way}
& \multicolumn{2}{c}{2-way}
& \multicolumn{2}{c}{3-way}
& \multicolumn{2}{c}{4-way} \\
\cmidrule(lr){2-3}\cmidrule(lr){4-5}\cmidrule(lr){6-7}
\cmidrule(lr){8-9}\cmidrule(lr){10-11}\cmidrule(lr){12-13}
\cmidrule(lr){14-15}\cmidrule(lr){16-17}\cmidrule(lr){18-19}
\cmidrule(lr){20-21}\cmidrule(lr){22-23}\cmidrule(lr){24-25}
\textbf{Explan. Condition}
  & 1s & 5s & 1s & 5s & 1s & 5s
  & 1s & 5s & 1s & 5s & 1s & 5s
  & 1s & 5s & 1s & 5s & 1s & 5s
  & 1s & 5s & 1s & 5s & 1s & 5s \\
\midrule
Classif.$^\dagger$
  & 100.0 & 100.0 & 100.0 & 100.0 & 100.0 & 100.0
  & 100.0 & 100.0 & 91.7 & 100.0 & 83.3 & 100.0
  & 100.0 & 100.0 & 100.0 & 100.0 & 83.3 & 100.0
  & 75.0 & 100.0 & 100.0 & 83.3 & 66.7 & 83.3 \\
NLE
  & 100.0 & 100.0 & 100.0 & 100.0 & 100.0 & 100.0
  & 100.0 & 100.0 & 66.7 & 100.0 & 83.3 & 91.7
  & 100.0 & 100.0 & 100.0 & 100.0 & 91.7 & 100.0
  & 83.3 & 100.0 & 100.0 & 83.3 & 66.7 & 91.7 \\
Features
  & 100.0 & 100.0 & 100.0 & 100.0 & 100.0 & 100.0
  & 100.0 & 100.0 & 75.0 & 100.0 & 83.3 & 100.0
  & 100.0 & 100.0 & 100.0 & 100.0 & 91.7 & 100.0
  & 58.3 & 100.0 & 91.7 & 83.3 & 58.3 & 91.7 \\
Feature-value pairs
  & 100.0 & 100.0 & 100.0 & 100.0 & 100.0 & 100.0
  & 100.0 & 100.0 & 83.3 & 100.0 & 91.7 & 100.0
  & 100.0 & 100.0 & 91.7 & 100.0 & 83.3 & 100.0
  & 66.7 & 100.0 & 100.0 & 83.3 & 91.7 & 75.0 \\
DL Axioms
  & 100.0 & 100.0 & 91.7 & 100.0 & 100.0 & 100.0
  & 100.0 & 100.0 & 75.0 & 100.0 & 83.3 & 83.3
  & 100.0 & 100.0 & 100.0 & 100.0 & 83.3 & 100.0
  & 75.0 & 100.0 & 83.3 & 83.3 & 75.0 & 83.3 \\
\bottomrule
\end{tabular}
\end{table*}

% ---- Table A3: Qwen3 VL 8B ----
\begin{table*}[htbp!]
\centering
\caption{Full FS-ICL classification accuracy breakdown for model Qwen3 VL 8B (\%). $N\in\{2,3,4\}$, $K\in\{1,5\}$, $Q{=}1$.
  $^\dagger$Classification-only baseline.}
\label{tab:full_qwen}
\scriptsize
\setlength{\tabcolsep}{2.5pt}
\renewcommand{\arraystretch}{1.1}
\begin{tabular}{@{}l
    cc cc cc
    cc cc cc
    cc cc cc
    cc cc cc@{}}
\toprule
& \multicolumn{6}{c}{\textbf{Flowers}}
& \multicolumn{6}{c}{\textbf{Oxford Pets}}
& \multicolumn{6}{c}{\textbf{CIFAR-10}}
& \multicolumn{6}{c}{\textbf{DTD}} \\
\cmidrule(lr){2-7}\cmidrule(lr){8-13}\cmidrule(lr){14-19}\cmidrule(lr){20-25}
& \multicolumn{2}{c}{2-way}
& \multicolumn{2}{c}{3-way}
& \multicolumn{2}{c}{4-way}
& \multicolumn{2}{c}{2-way}
& \multicolumn{2}{c}{3-way}
& \multicolumn{2}{c}{4-way}
& \multicolumn{2}{c}{2-way}
& \multicolumn{2}{c}{3-way}
& \multicolumn{2}{c}{4-way}
& \multicolumn{2}{c}{2-way}
& \multicolumn{2}{c}{3-way}
& \multicolumn{2}{c}{4-way} \\
\cmidrule(lr){2-3}\cmidrule(lr){4-5}\cmidrule(lr){6-7}
\cmidrule(lr){8-9}\cmidrule(lr){10-11}\cmidrule(lr){12-13}
\cmidrule(lr){14-15}\cmidrule(lr){16-17}\cmidrule(lr){18-19}
\cmidrule(lr){20-21}\cmidrule(lr){22-23}\cmidrule(lr){24-25}
\textbf{Explan. Condition}
  & 1s & 5s & 1s & 5s & 1s & 5s
  & 1s & 5s & 1s & 5s & 1s & 5s
  & 1s & 5s & 1s & 5s & 1s & 5s
  & 1s & 5s & 1s & 5s & 1s & 5s \\
\midrule
Classif.$^\dagger$
  & 100.0 & 100.0 & 100.0 & 100.0 & 100.0 & 100.0
  & 100.0 & 100.0 & 83.3 & 100.0 & 83.3 & 100.0
  & 100.0 & 100.0 & 91.7 & 100.0 & 83.3 & 100.0
  & 91.7 & 100.0 & 91.7 & 75.0 & 100.0 & 83.3 \\
NLE
  & 100.0 & 100.0 & 100.0 & 100.0 & 100.0 & 100.0
  & 100.0 & 100.0 & 66.7 & 100.0 & 83.3 & 100.0
  & 91.7 & 100.0 & 91.7 & 100.0 & 83.3 & 91.7
  & 83.3 & 100.0 & 83.3 & 75.0 & 100.0 & 75.0 \\
Features
  & 100.0 & 100.0 & 100.0 & 100.0 & 100.0 & 100.0
  & 100.0 & 100.0 & 75.0 & 100.0 & 83.3 & 91.7
  & 100.0 & 100.0 & 91.7 & 100.0 & 83.3 & 91.7
  & 83.3 & 100.0 & 100.0 & 83.3 & 91.7 & 75.0 \\
Feature-value pairs
  & 100.0 & 100.0 & 100.0 & 100.0 & 100.0 & 100.0
  & 100.0 & 100.0 & 58.3 & 100.0 & 83.3 & 91.7
  & 91.7 & 100.0 & 91.7 & 100.0 & 75.0 & 100.0
  & 91.7 & 91.7 & 91.7 & 83.3 & 91.7 & 75.0 \\
DL Axioms
  & 100.0 & 100.0 & 66.7 & 100.0 & 83.3 & 100.0
  & 100.0 & 100.0 & 75.0 & 100.0 & 66.7 & 66.7
  & 91.7 & 100.0 & 50.0 & 100.0 & 58.3 & 91.7
  & 75.0 & 75.0 & 75.0 & 83.3 & 66.7 & 66.7 \\
\bottomrule
\end{tabular}
\end{table*}

% ---- Table A4: LLaMA 4 Scout ----
\begin{table*}[htbp!]
\centering
\caption{Full FS-ICL classification accuracy breakdown for model LLaMA 4 Scout (\%). $N\in\{2,3,4\}$, $K\in\{1,5\}$, $Q{=}1$.
  $^\dagger$Classification-only baseline.}
\label{tab:full_llama}
\scriptsize
\setlength{\tabcolsep}{2.5pt}
\renewcommand{\arraystretch}{1.1}
\begin{tabular}{@{}l
    cc cc cc
    cc cc cc
    cc cc cc
    cc cc cc@{}}
\toprule
& \multicolumn{6}{c}{\textbf{Flowers}}
& \multicolumn{6}{c}{\textbf{Oxford Pets}}
& \multicolumn{6}{c}{\textbf{CIFAR-10}}
& \multicolumn{6}{c}{\textbf{DTD}} \\
\cmidrule(lr){2-7}\cmidrule(lr){8-13}\cmidrule(lr){14-19}\cmidrule(lr){20-25}
& \multicolumn{2}{c}{2-way}
& \multicolumn{2}{c}{3-way}
& \multicolumn{2}{c}{4-way}
& \multicolumn{2}{c}{2-way}
& \multicolumn{2}{c}{3-way}
& \multicolumn{2}{c}{4-way}
& \multicolumn{2}{c}{2-way}
& \multicolumn{2}{c}{3-way}
& \multicolumn{2}{c}{4-way}
& \multicolumn{2}{c}{2-way}
& \multicolumn{2}{c}{3-way}
& \multicolumn{2}{c}{4-way} \\
\cmidrule(lr){2-3}\cmidrule(lr){4-5}\cmidrule(lr){6-7}
\cmidrule(lr){8-9}\cmidrule(lr){10-11}\cmidrule(lr){12-13}
\cmidrule(lr){14-15}\cmidrule(lr){16-17}\cmidrule(lr){18-19}
\cmidrule(lr){20-21}\cmidrule(lr){22-23}\cmidrule(lr){24-25}
\textbf{Explan. Condition}
  & 1s & 5s & 1s & 5s & 1s & 5s
  & 1s & 5s & 1s & 5s & 1s & 5s
  & 1s & 5s & 1s & 5s & 1s & 5s
  & 1s & 5s & 1s & 5s & 1s & 5s \\
\midrule
Classif.$^\dagger$
  & 100.0 & 100.0 & 100.0 & 100.0 & 100.0 & 100.0
  & 100.0 & 100.0 & 66.7 & 100.0 & 83.3 & 100.0
  & 91.7 & 100.0 & 75.0 & 100.0 & 58.3 & 100.0
  & 66.7 & 100.0 & 75.0 & 75.0 & 66.7 & 75.0 \\
NLE
  & 100.0 & 100.0 & 100.0 & 100.0 & 100.0 & 100.0
  & 100.0 & 100.0 & 66.7 & 100.0 & 91.7 & 100.0
  & 91.7 & 100.0 & 75.0 & 100.0 & 83.3 & 100.0
  & 75.0 & 100.0 & 83.3 & 75.0 & 66.7 & 58.3 \\
Features
  & 100.0 & 100.0 & 100.0 & 100.0 & 100.0 & 100.0
  & 100.0 & 100.0 & 58.3 & 100.0 & 75.0 & 100.0
  & 91.7 & 100.0 & 83.3 & 100.0 & 66.7 & 100.0
  & 66.7 & 100.0 & 91.7 & 75.0 & 58.3 & 58.3 \\
Feature-value pairs
  & 100.0 & 100.0 & 91.7 & 100.0 & 100.0 & 100.0
  & 100.0 & 100.0 & 83.3 & 100.0 & 75.0 & 83.3
  & 83.3 & 100.0 & 66.7 & 91.7 & 83.3 & 83.3
  & 75.0 & 100.0 & 83.3 & 66.7 & 41.7 & 66.7 \\
DL Axioms
  & 100.0 & 100.0 & 100.0 & 100.0 & 75.0 & 100.0
  & 91.7 & 100.0 & 50.0 & 100.0 & 100.0 & 100.0
  & 100.0 & 100.0 & 66.7 & 91.7 & 66.7 & 91.7
  & 75.0 & 100.0 & 91.7 & 83.3 & 75.0 & 75.0 \\
\bottomrule
\end{tabular}
\end{table*}

\section{Detailed LLM-as-a-judge Results}
\label{app:judge_extra}

Tables~\ref{tab:acc_config_model}--\ref{tab:judge_dataset_metric}
report additional accuracy and judge score breakdowns.

\begin{table}[htbp!]
\centering
\caption{Mean accuracy (\%) by few-shot \icl configuration and model,
  aggregated over 4~datasets and 5~explanation types. $N\in\{2,3,4\}$, $K\in\{1,5\}$, $Q{=}1$.
  Values: mean~($\pm$SE). Best per row in bold.}
\label{tab:acc_config_model}
\small
\setlength{\tabcolsep}{3pt}
\renewcommand{\arraystretch}{1.1}
\begin{tabular}{@{}lcccc@{}}
\toprule
\textbf{Config} ($N$, $K$)
  & \textbf{Gem.~2.5F} & \textbf{Gem.~4} & \textbf{Qwen} & \textbf{LLaMA} \\
\midrule
$N{=}2$, $K{=}1$ & \textbf{95.8}~{\tiny(1.3)} & 92.9~{\tiny(1.7)} & 95.0~{\tiny(1.4)} & 90.4~{\tiny(1.9)} \\
$N{=}2$, $K{=}5$ & \textbf{100.0}~{\tiny(0.0)} & \textbf{100.0}~{\tiny(0.0)} & 98.3~{\tiny(0.8)} & \textbf{100.0}~{\tiny(0.0)} \\
$N{=}3$, $K{=}1$ & \textbf{97.1}~{\tiny(1.1)} & 92.5~{\tiny(1.7)} & 84.2~{\tiny(2.4)} & 80.4~{\tiny(2.6)} \\
$N{=}3$, $K{=}5$ & \textbf{100.0}~{\tiny(0.0)} & 95.8~{\tiny(1.3)} & 95.0~{\tiny(1.4)} & 92.9~{\tiny(1.7)} \\
$N{=}4$, $K{=}1$ & \textbf{90.4}~{\tiny(1.9)} & 85.8~{\tiny(2.3)} & 85.8~{\tiny(2.3)} & 78.3~{\tiny(2.7)} \\
$N{=}4$, $K{=}5$ & \textbf{96.2}~{\tiny(1.2)} & 95.0~{\tiny(1.4)} & 90.0~{\tiny(1.9)} & 89.2~{\tiny(2.0)} \\
\bottomrule
\end{tabular}
\end{table}

\begin{table}[htbp!]
\centering
\caption{Mean accuracy (\%) by model and dataset, aggregated over
  5~explanation types and 6~few-shot configurations:
  $5\times26=130$ observations per cell. $N\in\{2,3,4\}$, $K\in\{1,5\}$, $Q{=}1$.
  Values: mean~($\pm$SE). Best per column in bold.}
\label{tab:acc_model_dataset}
\small
\setlength{\tabcolsep}{3pt}
\renewcommand{\arraystretch}{1.1}
\begin{tabular}{@{}lcccc@{}}
\toprule
\textbf{Model}
  & \textbf{Flowers} & \textbf{Pets} & \textbf{CIFAR} & \textbf{DTD} \\
\midrule
Gemini 2.5 Flash & \textbf{100.0}~{\tiny(0.0)} & \textbf{98.9}~{\tiny(0.6)} & 95.8~{\tiny(1.1)} & \textbf{91.7}~{\tiny(1.5)} \\
Gemma 4 26B      & 99.7~{\tiny(0.3)} & 93.1~{\tiny(1.3)} & \textbf{97.5}~{\tiny(0.8)} & 84.4~{\tiny(1.9)} \\
Qwen3 VL 8B      & 98.3~{\tiny(0.7)} & 90.3~{\tiny(1.6)} & 91.7~{\tiny(1.5)} & 85.3~{\tiny(1.9)} \\
LLaMA 4 Scout    & 98.9~{\tiny(0.6)} & 90.8~{\tiny(1.5)} & 88.1~{\tiny(1.7)} & 76.4~{\tiny(2.2)} \\
\bottomrule
\end{tabular}
\end{table}

\begin{table*}[htbp!]
\centering
\caption{Best-performing explanation condition per model and dataset
  (highest mean accuracy averaged over $N\in\{2,3,4\}$, $K\in\{1,5\}$, $Q{=}1$).}
\label{tab:best_cond}
\small
\setlength{\tabcolsep}{4pt}
\renewcommand{\arraystretch}{1.1}
\begin{tabular}{@{}lcccccccc@{}}
\toprule
& \multicolumn{2}{c}{\textbf{Flowers}}
& \multicolumn{2}{c}{\textbf{Oxford Pets}}
& \multicolumn{2}{c}{\textbf{CIFAR-10}}
& \multicolumn{2}{c}{\textbf{DTD}} \\
\cmidrule(lr){2-3}\cmidrule(lr){4-5}\cmidrule(lr){6-7}\cmidrule(lr){8-9}
\textbf{Model}
  & \textbf{Best cond.} & \textbf{Acc.}
  & \textbf{Best cond.} & \textbf{Acc.}
  & \textbf{Best cond.} & \textbf{Acc.}
  & \textbf{Best cond.} & \textbf{Acc.} \\
\midrule
Gemini 2.5 Flash & Classification & 100.0 & Classification & 100.0 & Classification & 95.8 & Classification & 93.1 \\
Gemma 4 26B      & Classification & 100.0 & Classification & 95.8  & NLE            & 98.6 & NLE            & 87.5 \\
Qwen3 VL 8B      & Classification & 100.0 & Classification & 94.4  & Classification & 95.8 & Classification & 90.3 \\
LLaMA 4 Scout    & Classification & 100.0 & NLE            & 93.1  & NLE            & 91.7 & DL Axioms      & 86.1 \\
\bottomrule
\end{tabular}
\end{table*}

\begin{table*}[htbp!]
\centering
\caption{Mean LLM-as-a-judge score (1--5\,$\uparrow$) by model and metric,
  aggregated over 4~explanation conditions, 4~datasets, and 6~few-shot
  configurations: $4\times26=104$ observations per cell. $N\in\{2,3,4\}$, $K\in\{1,5\}$, $Q{=}1$.
  Values: mean~($\pm$SE). Best per XAI metric in \textbf{bold}.}
\label{tab:judge_model_metric}
\small
\setlength{\tabcolsep}{4pt}
\renewcommand{\arraystretch}{1.1}
\begin{tabular}{@{}lccccccccc@{}}
\toprule
\textbf{Model}
  & \textbf{TG} & \textbf{HF} & \textbf{CC} & \textbf{CP}
  & \textbf{Cn} & \textbf{S} & \textbf{LD} & \textbf{IF} & \textbf{LC} \\
\midrule
Gemini 2.5 Flash
  & \textbf{3.78}~{\tiny(.03)} & 4.61~{\tiny(.03)} & 4.57~{\tiny(.04)} & 4.79~{\tiny(.01)}
  & \textbf{4.95}~{\tiny(.01)} & \textbf{4.17}~{\tiny(.02)} & \textbf{4.18}~{\tiny(.03)} & 4.30~{\tiny(.03)} & 4.38~{\tiny(.03)} \\
Gemma 4 26B
  & 3.52~{\tiny(.03)} & \textbf{4.77}~{\tiny(.02)} & \textbf{4.65}~{\tiny(.03)} & \textbf{4.78}~{\tiny(.01)}
  & 4.88~{\tiny(.01)} & 3.90~{\tiny(.03)} & 3.88~{\tiny(.04)} & \textbf{4.38}~{\tiny(.03)} & \textbf{4.41}~{\tiny(.03)} \\
Qwen3 VL 8B
  & 3.01~{\tiny(.04)} & 4.54~{\tiny(.03)} & 4.47~{\tiny(.04)} & 4.58~{\tiny(.02)}
  & 4.96~{\tiny(.01)} & 3.21~{\tiny(.04)} & 3.21~{\tiny(.04)} & 4.20~{\tiny(.03)} & 4.44~{\tiny(.03)} \\
LLaMA 4 Scout
  & 2.95~{\tiny(.03)} & 4.52~{\tiny(.03)} & 4.24~{\tiny(.04)} & 4.68~{\tiny(.02)}
  & 4.88~{\tiny(.01)} & 3.25~{\tiny(.03)} & 3.05~{\tiny(.04)} & 4.13~{\tiny(.03)} & 4.19~{\tiny(.04)} \\
\bottomrule
\end{tabular}
\end{table*}

\begin{table*}[htbp!]
\centering
\caption{Mean LLM-as-a-judge score (1--5\,$\uparrow$) by dataset and metric,
  aggregated over 4~models, 4~explanation conditions, and 6~few-shot
  configurations: $4\times26=104$ observations per cell. $N\in\{2,3,4\}$, $K\in\{1,5\}$, $Q{=}1$. 
  Values: mean~($\pm$SE). \textbf{Bold} = best per XAI metric.}
\label{tab:judge_dataset_metric}
\small
\setlength{\tabcolsep}{4pt}
\renewcommand{\arraystretch}{1.1}
\begin{tabular}{@{}lccccccccc@{}}
\toprule
\textbf{Dataset}
  & \textbf{TG} & \textbf{HF} & \textbf{CC} & \textbf{CP}
  & \textbf{Cn} & \textbf{S} & \textbf{LD} & \textbf{IF} & \textbf{LC} \\
\midrule
Flowers
  & \textbf{3.38}~{\tiny(.03)} & \textbf{4.64}~{\tiny(.02)} & 4.36~{\tiny(.04)} & 4.72~{\tiny(.02)}
  & \textbf{4.96}~{\tiny(.01)} & \textbf{3.79}~{\tiny(.03)} & \textbf{3.64}~{\tiny(.04)} & \textbf{4.36}~{\tiny(.03)} & \textbf{4.50}~{\tiny(.03)} \\
Oxford Pets
  & 3.20~{\tiny(.03)} & 4.71~{\tiny(.02)} & \textbf{4.76}~{\tiny(.03)} & \textbf{4.74}~{\tiny(.02)}
  & 4.96~{\tiny(.01)} & 3.66~{\tiny(.03)} & 3.55~{\tiny(.04)} & 4.35~{\tiny(.03)} & 4.41~{\tiny(.03)} \\
CIFAR-10
  & 3.21~{\tiny(.04)} & 4.56~{\tiny(.03)} & 4.45~{\tiny(.04)} & 4.70~{\tiny(.02)}
  & 4.92~{\tiny(.01)} & 3.37~{\tiny(.03)} & 3.60~{\tiny(.04)} & 4.16~{\tiny(.03)} & 4.27~{\tiny(.04)} \\
DTD
  & 3.46~{\tiny(.04)} & 4.53~{\tiny(.03)} & 4.36~{\tiny(.04)} & 4.68~{\tiny(.02)}
  & 4.83~{\tiny(.02)} & 3.71~{\tiny(.03)} & 3.53~{\tiny(.04)} & 4.14~{\tiny(.03)} & 4.23~{\tiny(.04)} \\
\bottomrule
\end{tabular}
\end{table*}

\begin{figure}[htbp!]
\centering
\includegraphics[width=\linewidth]{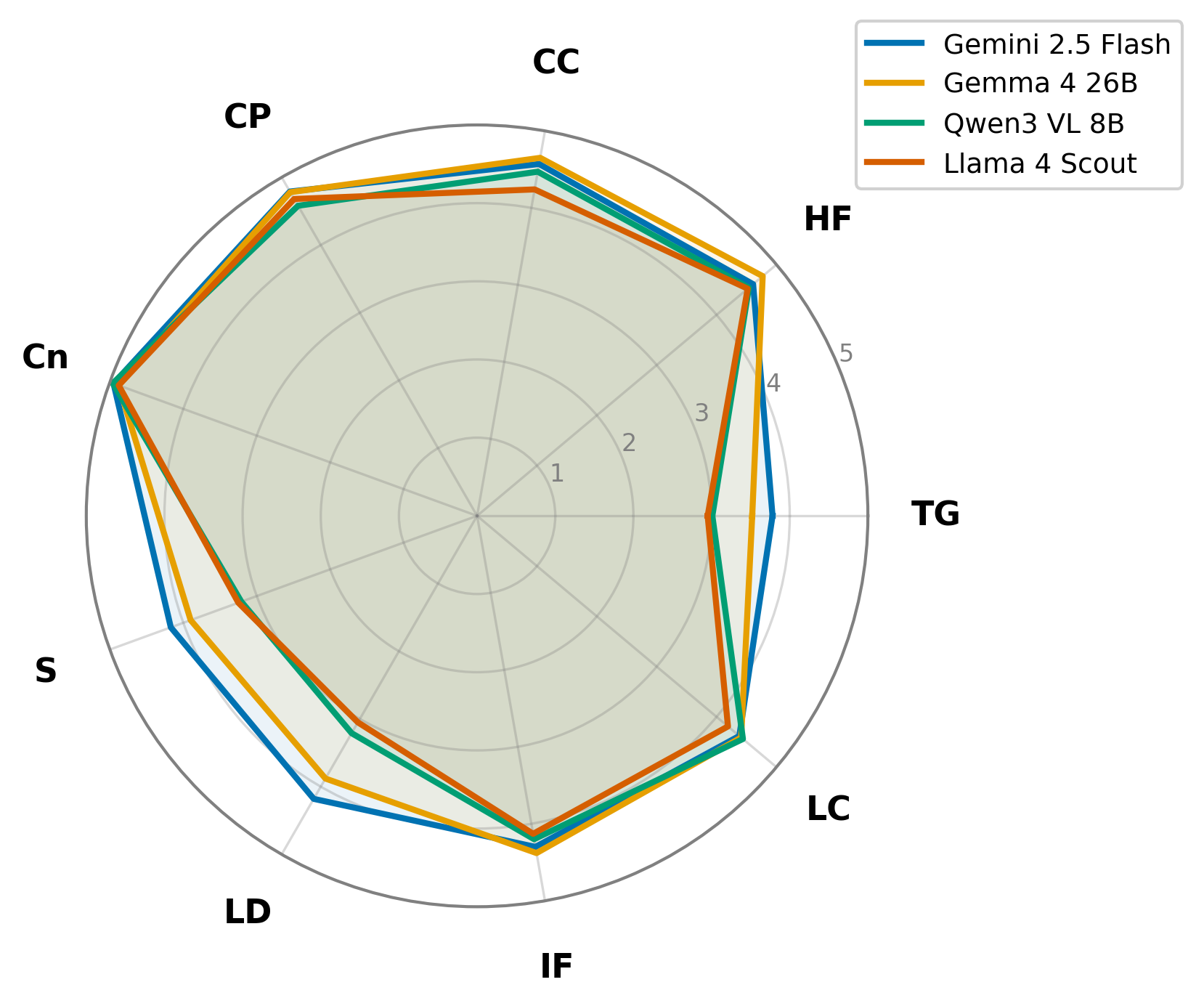}
\caption{LLM-as-a-judge scores by model, averaged over all conditions
  and datasets. Gemini 2.5 Flash maintains the largest polygon
  on TG, S, and LD; Qwen3 VL 8B and LLaMA 4 Scout show the
  most pronounced collapses. The metrics are: Textual Groundedness (TG), Hallucination Free (HF), Concept Counting (CC), Comprehensibility (CP), Conciseness (Cn), Specificity (S), Local Discriminativeness (LD), Instruction Following (IF), and Logical Coherence (LC)}
\label{fig:radar_model}
\end{figure}

%\cnat{ seria ideal si podemos poner acronimos de metricas en el plot's caption}
% ---- Table B5: Judge Breakdown Gemini 2.5 Flash ----
\begin{table*}[htbp!]
\centering
\caption{Full LLM-as-a-judge score breakdown (1--5\,$\uparrow$) for model ( Gemini 2.5 Flash). $N\in\{2,3,4\}$, $K\in\{1,5\}$, $Q{=}1$.
  Observations per cell vary by $N$: $N{=}2 \to 6$ observations, $N{=}3 \to 4$ observations, $N{=}4 \to 3$ observations.
  Values: mean~($\pm$SE). \textbf{Bold} indicates best score per XAI metric.}
\label{tab:judge_gemini}
\small
\setlength{\tabcolsep}{4pt}
\renewcommand{\arraystretch}{1.1}
\begin{tabular}{@{}lccccccccc@{}}
\toprule
\textbf{Explanation Condition}
  & \textbf{TG} & \textbf{HF} & \textbf{CC} & \textbf{CP}
  & \textbf{Cn} & \textbf{S} & \textbf{LD} & \textbf{IF} & \textbf{LC} \\
\midrule
NLE
  & 3.76~{\tiny(.04)} & 4.49~{\tiny(.05)} & \textbf{4.77}~{\tiny(.05)} & \textbf{4.99}~{\tiny(.01)}
  & 4.90~{\tiny(.02)} & 4.02~{\tiny(.04)} & 4.17~{\tiny(.06)} & 4.67~{\tiny(.03)} & \textbf{4.96}~{\tiny(.02)} \\
Features
  & 3.94~{\tiny(.05)} & 4.77~{\tiny(.04)} & 4.63~{\tiny(.06)} & 4.97~{\tiny(.01)}
  & 4.97~{\tiny(.01)} & 4.08~{\tiny(.05)} & 4.15~{\tiny(.07)} & \textbf{4.72}~{\tiny(.04)} & 4.93~{\tiny(.02)} \\
Feature-value pairs
  & \textbf{4.24}~{\tiny(.03)} & \textbf{4.82}~{\tiny(.03)} & 4.51~{\tiny(.07)} & 4.97~{\tiny(.01)}
  & \textbf{4.98}~{\tiny(.01)} & \textbf{4.61}~{\tiny(.03)} & \textbf{4.39}~{\tiny(.06)} & 4.70~{\tiny(.04)} & 4.76~{\tiny(.04)} \\
DL Axioms
  & 3.16~{\tiny(.06)} & 4.38~{\tiny(.07)} & 4.39~{\tiny(.09)} & 4.20~{\tiny(.03)}
  & 4.94~{\tiny(.02)} & 3.98~{\tiny(.06)} & 4.01~{\tiny(.07)} & 3.12~{\tiny(.05)} & 2.86~{\tiny(.07)} \\
\bottomrule
\end{tabular}
\end{table*}

%\cnat{como se obtienen las flechitas y 4 reps, a partir de los N? son repetitions? intentar aclararlo en cada caption.  no entiendo N=2 flecha 6 reps}

% ---- Table B6: Judge Breakdown Gemma 4 26B ----
\begin{table*}[htbp!]
\centering
\caption{Full LLM-as-a-judge score breakdown (1--5\,$\uparrow$) for model (Gemma 4 26B). $N\in\{2,3,4\}$, $K\in\{1,5\}$, $Q{=}1$.
  Observations per cell vary by $N$: $N{=}2 \to 6$ observations, $N{=}3 \to 4$ observations, $N{=}4 \to 3$ observations.
  Values: mean~($\pm$SE). \textbf{Bold} indicates best score per XAI metric.}
\label{tab:judge_gemma}
\small
\setlength{\tabcolsep}{4pt}
\renewcommand{\arraystretch}{1.1}
\begin{tabular}{@{}lccccccccc@{}}
\toprule
\textbf{Explanation Condition}
  & \textbf{TG} & \textbf{HF} & \textbf{CC} & \textbf{CP}
  & \textbf{Cn} & \textbf{S} & \textbf{LD} & \textbf{IF} & \textbf{LC} \\
\midrule
NLE
  & 3.48~{\tiny(.05)} & 4.61~{\tiny(.05)} & 4.63~{\tiny(.07)} & 4.93~{\tiny(.02)}
  & 4.83~{\tiny(.04)} & 3.61~{\tiny(.05)} & 3.69~{\tiny(.08)} & 4.82~{\tiny(.03)} & 4.81~{\tiny(.03)} \\
Features
  & 3.71~{\tiny(.05)} & \textbf{4.91}~{\tiny(.02)} & \textbf{4.88}~{\tiny(.04)} & \textbf{5.00}~{\tiny(.00)}
  & \textbf{4.99}~{\tiny(.01)} & 3.92~{\tiny(.05)} & 3.74~{\tiny(.08)} & \textbf{4.95}~{\tiny(.01)} & \textbf{4.93}~{\tiny(.02)} \\
Feature-value pairs
  & \textbf{3.92}~{\tiny(.04)} & 4.88~{\tiny(.03)} & 4.61~{\tiny(.07)} & 4.90~{\tiny(.02)}
  & 4.86~{\tiny(.03)} & \textbf{4.29}~{\tiny(.04)} & \textbf{4.11}~{\tiny(.07)} & 4.56~{\tiny(.05)} & 4.69~{\tiny(.04)} \\
DL Axioms
  & 2.96~{\tiny(.06)} & 4.66~{\tiny(.05)} & 4.50~{\tiny(.08)} & 4.30~{\tiny(.04)}
  & 4.86~{\tiny(.03)} & 3.79~{\tiny(.05)} & 3.99~{\tiny(.07)} & 3.17~{\tiny(.05)} & 3.20~{\tiny(.08)} \\
\bottomrule
\end{tabular}
\end{table*}

%\cnat{intentar ser mas specifico, per column no, per XAI metric, si el tamaño de las tablas lo permite para ser coherente con definiciones, renombrar a feature-value que queda regular en singular a feature-value pairs}

% ---- Table B7: Judge Breakdown Qwen3 VL 8B ----
\begin{table*}[htbp!]
\centering
\caption{Full LLM-as-a-judge score breakdown (1--5\,$\uparrow$) for model (Qwen3 VL 8B). $N\in\{2,3,4\}$, $K\in\{1,5\}$, $Q{=}1$.
  Observations per cell vary by $N$: $N{=}2 \to 6$ observations, $N{=}3 \to 4$ observations, $N{=}4 \to 3$ observations.
  Values: mean~($\pm$SE). \textbf{Bold} indicates best score per XAI metric.} 
\label{tab:judge_qwen}
\small
\setlength{\tabcolsep}{4pt}
\renewcommand{\arraystretch}{1.1}
\begin{tabular}{@{}lccccccccc@{}}
\toprule
\textbf{Explanation Condition}
  & \textbf{TG} & \textbf{HF} & \textbf{CC} & \textbf{CP}
  & \textbf{Cn} & \textbf{S} & \textbf{LD} & \textbf{IF} & \textbf{LC} \\
\midrule
NLE
  & \textbf{3.76}~{\tiny(.04)} & 4.35~{\tiny(.06)} & \textbf{4.77}~{\tiny(.05)} & 4.96~{\tiny(.01)}
  & 4.84~{\tiny(.02)} & 3.85~{\tiny(.04)} & 3.75~{\tiny(.07)} & 4.64~{\tiny(.04)} & 4.85~{\tiny(.03)} \\
Features
  & 3.55~{\tiny(.06)} & \textbf{4.83}~{\tiny(.03)} & 4.76~{\tiny(.05)} & \textbf{4.99}~{\tiny(.01)}
  & 4.99~{\tiny(.01)} & 3.79~{\tiny(.05)} & 3.42~{\tiny(.08)} & \textbf{4.90}~{\tiny(.02)} & 4.89~{\tiny(.03)} \\
Feature-value pairs
  & 3.66~{\tiny(.05)} & 4.81~{\tiny(.03)} & 4.38~{\tiny(.08)} & \textbf{4.99}~{\tiny(.01)}
  & \textbf{5.00}~{\tiny(.00)} & \textbf{4.08}~{\tiny(.04)} & \textbf{3.97}~{\tiny(.07)} & 4.24~{\tiny(.06)} & \textbf{4.92}~{\tiny(.02)} \\
DL Axioms
  & 1.06~{\tiny(.02)} & 4.16~{\tiny(.07)} & 3.97~{\tiny(.10)} & 3.39~{\tiny(.04)}
  & \textbf{5.00}~{\tiny(.00)} & 1.09~{\tiny(.03)} & 1.69~{\tiny(.07)} & 3.00~{\tiny(.06)} & 3.10~{\tiny(.09)} \\
\bottomrule
\end{tabular}
\end{table*}

%\cnat{escribir bien LLaMA en todas sus apariciones. El caption de las tablas 19 y analogas podria tener la flechita para arriba tras el 1-5, y añadir for model (model name. ). Incluir los k y sus valores para que no de lugar a duda las flechas, si es FSL, k,q,n debe quedar seimpre claro en todas las captions. }
% ---- Table B8: Judge Breakdown LLaMA 4 Scout ----
\begin{table*}[htbp!]
\centering
\caption{Full LLM-as-a-judge score breakdown (1--5\,$\uparrow$) for model (LLaMA 4 Scout). $N\in\{2,3,4\}$, $K\in\{1,5\}$, $Q{=}1$.
  Observations per cell vary by $N$: $N{=}2 \to 6$ observations, $N{=}3 \to 4$ observations, $N{=}4 \to 3$ observations.
  Values: mean~($\pm$SE). \textbf{Bold} indicates best score per XAI metric.}
\label{tab:judge_llama}
\small
\setlength{\tabcolsep}{4pt}
\renewcommand{\arraystretch}{1.1}
\begin{tabular}{@{}lccccccccc@{}}
\toprule
\textbf{Explanation Condition}
  & \textbf{TG} & \textbf{HF} & \textbf{CC} & \textbf{CP}
  & \textbf{Cn} & \textbf{S} & \textbf{LD} & \textbf{IF} & \textbf{LC} \\
\midrule
NLE
  & \textbf{3.48}~{\tiny(.05)} & 4.40~{\tiny(.06)} & \textbf{4.56}~{\tiny(.07)} & 4.92~{\tiny(.02)}
  & 4.67~{\tiny(.04)} & 3.43~{\tiny(.05)} & 3.15~{\tiny(.08)} & 4.68~{\tiny(.04)} & 4.74~{\tiny(.04)} \\
Features
  & 3.30~{\tiny(.06)} & \textbf{4.72}~{\tiny(.04)} & 4.47~{\tiny(.08)} & \textbf{4.99}~{\tiny(.01)}
  & 4.93~{\tiny(.02)} & 3.44~{\tiny(.05)} & 3.17~{\tiny(.08)} & \textbf{4.73}~{\tiny(.04)} & \textbf{4.79}~{\tiny(.04)} \\
Feature-value pairs
  & 2.97~{\tiny(.06)} & 4.57~{\tiny(.05)} & 4.00~{\tiny(.10)} & 4.83~{\tiny(.03)}
  & \textbf{4.97}~{\tiny(.01)} & \textbf{3.57}~{\tiny(.05)} & \textbf{3.19}~{\tiny(.08)} & 4.22~{\tiny(.06)} & 4.49~{\tiny(.05)} \\
DL Axioms
  & 2.05~{\tiny(.07)} & 4.40~{\tiny(.07)} & 3.94~{\tiny(.10)} & 4.00~{\tiny(.04)}
  & 4.94~{\tiny(.03)} & 2.57~{\tiny(.08)} & 2.69~{\tiny(.09)} & 2.91~{\tiny(.05)} & 2.72~{\tiny(.07)} \\
\bottomrule
\end{tabular}
\end{table*}

\section{Effect of $N$ and $K$ on Accuracy and Explanation Quality}
\label{app:nk_analysis}

\begin{figure*}[htbp!]
\centering
\includegraphics[width=\linewidth]{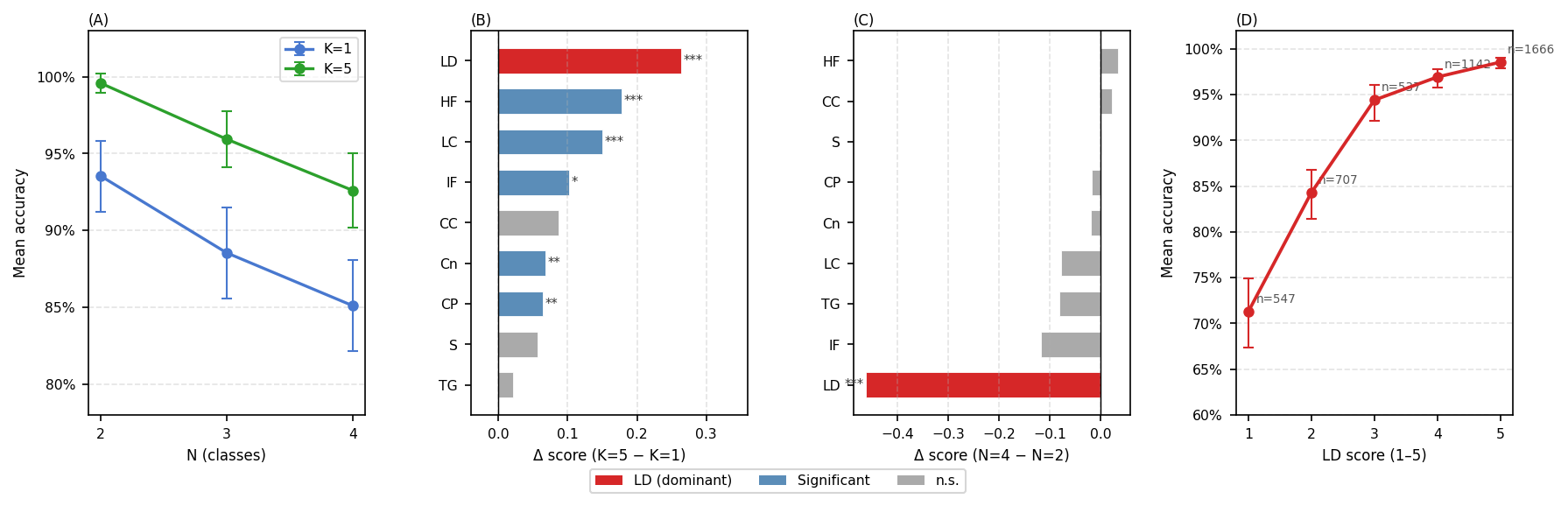}
\caption{%
  (A)~Mean classification accuracy vs.\ number of classes $N$, by support
  shots $K$. K=5 raises accuracy by $+7.0$\,pp on average
  ($p{=}2.0{\times}10^{-13}$, pooled Wilcoxon); accuracy falls
  monotonically with $N$ at both $K$ levels
  (Friedman $\chi^2{=}75.2$, $p{=}4.7{\times}10^{-17}$).
  Error bars: 95\% CI\@.
  (B)~Effect of $K$ on each XAI metric ($\Delta = \text{K=5} - \text{K=1}$,
  pooled across $N$; Bonferroni ${\times}40$).
  Local Discriminativeness shows the largest gain
  ($\Delta{=}{+}0.26$, $p_\text{Bonf}{=}1.5{\times}10^{-6}$);
  TG, CC, and S are unaffected.
  (C)~Effect of $N$ on each XAI metric ($\Delta = \text{N=4} - \text{N=2}$,
  pooled across $K$; Bonferroni ${\times}30$).
  Only Local Discriminativeness shows a significant N-trend
  ($\chi^2{=}51.5$, $p_\text{Bonf}{=}1.9{\times}10^{-10}$).
  (D)~Mean classification accuracy as a function of LD score
  (pooled across all models, datasets, and conditions E2--E5;
  $n$ = trials per bin; error bars: 95\% CI).
  Significance markers: {*}~$p_\text{Bonf}{<}0.05$,
  {**}~$p_\text{Bonf}{<}0.01$,
  {***}~$p_\text{Bonf}{<}0.001$.
  Metric abbreviations as in Table~\ref{tab:metrics}.}
\label{fig:nk_analysis}
\end{figure*}

This section provides a brief analysis of how the number of classes ($N$) and support examples ($K$) influence both accuracy and explanation quality. The main trends are summarised in Figure~\ref{fig:nk_analysis}.

% ============================================================
% NEW APPENDIX G
% (\section{Effect of $N$ and $K$...} / \label{app:nk_analysis})
% ============================================================
 
\section{Ablation Study: Effect of Support-Image Access on Judge Scores}
\label{app:judge_validation}
  
The judge pipeline described in Section~\ref{sec:evaluation} evaluates
explanations using only the query image, without access to the few-shot
support images seen by the classifier.
Beyond validating this specific design choice, this experiment reveals
which evaluation metrics are inherently sensitive to visual access and
which are not.
All results reported in Section~\ref{sec:results} are based exclusively
on the original GPT-QO judge and are unaffected by the limitations of
the Qwen judge described below; this appendix is a validation of the
original judge design, not a replacement or extension of those results.

To isolate the effect of support-image access, we ran a second judge
(\texttt{Qwen3-VL-32B-Thinking}) on the same explanations under two
conditions.
\textbf{Qwen-QueryOnly} (Qwen-QO) replicates the original setup:
query image only, no support images.
\textbf{Qwen-QueryAndSupport} (Qwen-QAS) additionally provides the
full set of labelled support images.
The original judge (\texttt{gpt-5-thinking-mini}) is labelled
\textbf{GPT-QueryOnly} (GPT-QO).
 
This design yields two distinct comparisons:
(i)~\emph{Qwen-QO vs.\ Qwen-QAS} isolates the effect of support-image
access within a fixed model;
(ii)~\emph{GPT-QO vs.\ Qwen-QO} probes between-model calibration under
identical conditions.
All three judges scored the same 3{,}504 trials.
The DL\,Axioms prompt type contributes fewer trials
(497 vs.\ $\approx$1{,}000; Table~\ref{tab:judge_trial_counts}) due to
parse errors by Qwen-QO on structured outputs (the judge model
exhausted its token budget while reasoning over the formal DL syntax
before producing the scoring XML, leaving those trials without a valid
score).
We report results both including and excluding this subset to verify
that conclusions do not depend on this incomplete subset.

It should be noted that the 497 scored axioms trials are not a random
sample of the full axioms set: they are most likely the trials whose
classifier outputs were shorter or structurally simpler, since more
elaborate DL\,Axiom outputs tend to exhaust the judge's token budget
before the scoring XML is produced.
This introduces a selection bias --- the correlations reported for the
axioms subset may be overestimated relative to what would be observed
on the full set.
For this reason, the main conclusions of this appendix rest on the
three prompt types where coverage is complete (Features, NLE,
Rule-based; $n = 3{,}007$ after removing axioms trials), and axioms
results are reported as additional evidence with this limitation
explicit.
Addressing this limitation would require either a non-reasoning judge
model (which foregoes the chain-of-thought transparency) or a dedicated
formal verifier for OWL\,2 outputs; both are left for future work given
the computational constraints of this study.
 
\begin{table}[htbp!]
\centering
\caption{Trial counts at the intersection of all three judge conditions
  by prompt type. The reduced count for DL\,Axioms is due to Qwen-QO
  parse errors. Individual-judge counts reflect all parseable outputs
  before intersection.}
\label{tab:judge_trial_counts}
\small
\setlength{\tabcolsep}{4pt}
\renewcommand{\arraystretch}{1.1}
\begin{tabular}{@{}lrrrr@{}}
\toprule
\textbf{Prompt type} & \textbf{Intersect.} & \textbf{GPT-QO} & \textbf{Qwen-QO} & \textbf{Qwen-QAS} \\
\midrule
DL Axioms  & 497     & 1{,}146 & 588     & 985     \\
Features   & 999     & 1{,}152 & 1{,}062 & 1{,}084 \\
NLE        & 1{,}030 & 1{,}151 & 1{,}075 & 1{,}101 \\
Rule-based & 978     & 1{,}150 & 1{,}050 & 1{,}069 \\
\midrule
\textbf{Total} & \textbf{3{,}504} & & & \\
\bottomrule
\end{tabular}
\end{table}
 
\subsection*{Results}
 
Table~\ref{tab:judge_validation} reports Spearman $\rho$ and mean score
differences for all nine metrics, Bonferroni-corrected over
27~simultaneous tests ($9~\text{metrics} \times 3~\text{pairs}$).
 
\begin{table*}[htbp!]
\centering
\caption{Cross-judge validation: Spearman $\rho$ [95\,\% CI] between
  judge conditions and mean score difference (GPT-QO $-$ Qwen-QAS) for
  all nine XAI metrics ($n = 3{,}504$ trials; all $p_{\mathrm{Bonf}} <
  0.001$).
  \textbf{Bold} highlights the Local Discriminativeness row.
  The Qwen-QO vs.\ Qwen-QAS column isolates the effect of support-image
  access within a fixed model.}
\label{tab:judge_validation}
\small
\setlength{\tabcolsep}{4pt}
\renewcommand{\arraystretch}{1.1}
\begin{tabular}{@{}l
    >{\centering\arraybackslash}p{2.8cm}
    >{\centering\arraybackslash}p{2.8cm}
    >{\centering\arraybackslash}p{2.8cm}
    >{\centering\arraybackslash}p{1.4cm}@{}}
\toprule
\textbf{Explainability Metric}
  & \textbf{GPT-QO vs.\ Qwen-QO} $\rho$ [CI]
  & \textbf{GPT-QO vs.\ Qwen-QAS} $\rho$ [CI]
  & \textbf{Qwen-QO vs.\ Qwen-QAS} $\rho$ [CI]
  & \textbf{Mean diff.\ GPT-QO$-$QAS} \\
\midrule
Comprehensibility    & 0.691 [0.655,\,0.725] & 0.733 [0.699,\,0.763] & 0.827 [0.801,\,0.853] & $+0.193$ \\
Specificity          & 0.549 [0.523,\,0.573] & 0.562 [0.536,\,0.586] & 0.620 [0.591,\,0.648] & $-0.478$ \\
\textbf{Local Discriminativeness}
  & \textbf{0.536 [0.509,\,0.562]}
  & \textbf{0.531 [0.502,\,0.558]}
  & \textbf{0.627 [0.598,\,0.653]}
  & $\mathbf{-0.421}$ \\
\midrule
Textual Groundedness & 0.456 [0.426,\,0.484] & 0.494 [0.468,\,0.521] & 0.562 [0.536,\,0.590] & $-0.339$ \\
Hallucination Free   & 0.339 [0.298,\,0.377] & 0.278 [0.237,\,0.319] & 0.427 [0.388,\,0.466] & $-0.031$ \\
\midrule
Conciseness          & 0.445 [0.391,\,0.497] & 0.361 [0.304,\,0.419] & 0.316 [0.266,\,0.364] & $+0.090$ \\
Instruction Following & 0.246 [0.207,\,0.283] & 0.315 [0.278,\,0.351] & 0.339 [0.286,\,0.390] & $-0.423$ \\
Logical Coherence    & 0.411 [0.370,\,0.450] & 0.381 [0.340,\,0.421] & 0.496 [0.451,\,0.538] & $-0.235$ \\
\midrule
Concept Counting     & 0.169 [0.127,\,0.212] & 0.170 [0.127,\,0.214] & 0.523 [0.477,\,0.566] & $-0.153$ \\
\bottomrule
\end{tabular}
\end{table*}
 
\subsection*{Interpretation}
 
The nine metrics fall into three groups according to how the Qwen-QO
vs.\ Qwen-QAS comparison (the clean within-model test) behaves.
 
\textbf{Metrics robust to visual access (Comprehensibility, Specificity,
Local Discriminativeness).}
These three metrics show the highest within-model agreement across
conditions ($\rho = 0.62$--$0.83$) and negligible mean score shifts
($|\Delta| \leq 0.06$ for the Qwen pair).
This pattern makes conceptual sense: Comprehensibility and Specificity
evaluate properties of the text itself that a judge can assess from the
explanation alone, regardless of what the support images look like.
Local Discriminativeness is notable in this group: despite requiring
class-contrastive reasoning, the ranking of explanations on LD is
essentially unaffected by whether the judge sees the support set
($\rho = 0.627$, mean diff $= -0.049$).
This supports the design rationale that LD can be reliably assessed
through the judge's prior class knowledge combined with the query image.
The GPT-QO vs.\ Qwen-QAS correlation for LD ($\rho = 0.531$) confirms
that even across different model families the ranking order is
preserved, with only a constant calibration offset of $-0.421$ points
that is uniform across prompt types.
 
\textbf{Metrics sensitive to visual access (Textual Groundedness,
Hallucination Free).}
Both metrics show notably lower ranking agreement between the two Qwen
conditions --- that is, lower Qwen-QO vs.\ Qwen-QAS correlation ---
($\rho = 0.43$--$0.56$) than the three metrics in the robust group
(Comprehensibility, Specificity, and Local Discriminativeness,
$\rho = 0.62$--$0.83$).
This is precisely what their definitions predict: Textual Groundedness
measures whether the explanation covers all salient image concepts, and
Hallucination Free measures whether every claim is visually verifiable
--- both are judgments that require seeing the images.
When the classifier generates an explanation that references visual
patterns from the support set (e.g.\ ``unlike the other examples, this
flower has broader petals''), a query-only judge cannot verify whether
that comparison is accurate, because it never saw those support images.
A judge with access to the full support set can, and will sometimes
score the same explanation differently as a result.
The lower Qwen-QO vs.\ Qwen-QAS agreement for these two metrics
therefore reflects a genuine limitation of query-only evaluation for
TG and HF specifically: the judge is missing part of the visual
evidence these metrics are defined against.
This is a known cost of our compute-constrained design and points
towards future evaluation pipelines that provide the full visual
context to the judge for these two metrics in particular.
 
\textbf{Intermediate metrics: form-dependent but context-sensitive
(Conciseness, Instruction Following, Logical Coherence).}
These three metrics might be expected to be entirely image-independent
because they evaluate structural or linguistic properties of the
explanation text.
Indeed, their mean score shifts between Qwen-QO and Qwen-QAS are
negligible ($|\Delta| < 0.06$).
However, their $\rho$ values are substantially lower than those of the
robust group (Comprehensibility, Specificity, and Local
Discriminativeness; $\rho = 0.62$--$0.83$), reaching only 0.32--0.50,
and this gap is largely driven by the DL\,Axioms subset.
Excluding axioms, the within-model correlations for Conciseness,
Instruction Following, and Logical Coherence recover considerably
($\rho = 0.37$, $0.30$, $0.48$ respectively), and cross-model
agreement for Logical Coherence approaches zero mean difference.
This suggests the lower correlations in the 3{,}504-trial set
(which includes the 497 DL\,Axiom trials) reflect the difficulty of
scoring those outputs consistently, rather than genuine sensitivity to
visual access per se.
When the classifier produces a malformed or partial OWL\,2 axiom,
metrics like Instruction Following and Logical Coherence become
ambiguous to score: the output is neither well-formed natural language
nor valid formal syntax, so any LLM judge that has not been specifically
calibrated on that format will make inconsistent judgments, and the
two Qwen conditions, despite sharing the same model, will sometimes
disagree for that reason alone.
Rigorously evaluating DL\,Axiom outputs would require formal
verification with a dedicated OWL\,2 reasoner
(e.g.\ OntoAligner~\citep{babaei2025ontoaligner} or similar tools),
which could check structural validity and logical consistency directly
rather than delegating that judgment to an LLM; we leave this for
future work.
 
\textbf{Concept Counting.}
Concept Counting stands apart: its within-model agreement is
moderate ($\rho = 0.52$) but its cross-model agreement is very weak
($\rho \approx 0.17$), the lowest of all nine metrics.
This pattern indicates that GPT and Qwen apply fundamentally different
interpretations of what constitutes a countable concept in an
explanation, a model-level disagreement that is independent of image
access.
We advise treating Concept Counting results with additional caution.
Unlike Local Discriminativeness --- where GPT and Qwen produce
different absolute scores but preserve a similar ranking order
($\rho = 0.53$) --- for Concept Counting not even the ranking is
preserved across models ($\rho \approx 0.17$).
This means the discrepancy is not a matter of calibration that cancels
in relative comparisons: the two models are to some extent measuring
different things.
Consequently, if this experiment were reproduced with a different judge
model, the Concept Counting scores would likely not be comparable to
ours --- not because the explanations changed, but because a different
judge may count concepts differently.
 
\subsection*{Summary and Implications}

Local Discriminativeness, the metric most central to the paper's
findings, falls in the robust group.
Two independent pieces of evidence support this.
First, within the same model, providing or withholding the support
images produces no meaningful reordering of explanations on LD
($\rho = 0.627$, mean score shift $= -0.049$ points on a 1--5 scale):
a judge that has never seen the support set reaches essentially the
same LD ranking as one that has.
Second, even across two different model families with different visual
access --- GPT-QO and Qwen-QAS --- the LD ranking is preserved
($\rho = 0.531$), with only a constant calibration offset of
$-0.421$ points that is uniform across all four prompt types
(NLE, Features, Rule-based, DL\,Axioms) and therefore cancels
entirely in the between-condition comparisons on which the paper's
conclusions rest.
In other words, the finding that LD is the strongest predictor of
classification accuracy, and that it responds specifically to $K$ and
$N$ manipulations, holds regardless of which judge model is used and
regardless of whether that judge sees the support images.
 
A natural concern is whether a judge without support images can
reliably assess a metric that requires class-contrastive reasoning.
The answer, empirically, is yes for LD: the judge's prior knowledge of
the candidate classes --- which are drawn from standard benchmarks
(CIFAR-10, DTD, Oxford Flowers, Oxford-IIIT Pets) --- is sufficient to
assess whether an explanation highlights features that separate the
predicted class from the alternatives.
The metrics that genuinely require full visual access are Textual
Groundedness and Hallucination Free, and this is theoretically expected:
both are defined in terms of verifying claims against the images, so a
judge that sees more images will score them differently.
This is a scoping limitation we acknowledge and address in future work,
not a threat to the validity of the LD-based conclusions.

% ============================================================
\section{Human Validation of the LLM-as-a-Judge Pipeline}
\label{app:human_validation}

To complement the automated ablation study in Appendix~\ref{app:judge_validation},
we conducted an independent human evaluation in which three researchers
manually scored a stratified sample of explanations using the same
rubrics provided to the judge model.
This appendix describes the annotation interface, the sampling protocol,
the statistical methods used, and the results of the human--judge
alignment analysis.

% ---- H.1 ----
\subsection*{Annotation Interface}

To collect human judgements, we developed a local web application
that presents each trial to the annotator in a three-column layout:
the query image with the predicted class label (highlighted in green
if correct, red if incorrect) on the left; the full explanation
generated by the classifier model, together with the labelled support
images in a collapsible panel, in the centre; and a scoring form on
the right, where each of the nine XAI metrics is rated via radio
buttons (1--5), with a per-metric expandable rubric available at all
times via an info button (\textbf{i}).
Annotations are written to disk immediately upon clicking
\emph{Save \& Next}, so that progress is preserved across sessions.
The application is included in the public repository.

\begin{figure*}[htbp!]
\centering
\includegraphics[width=\linewidth]{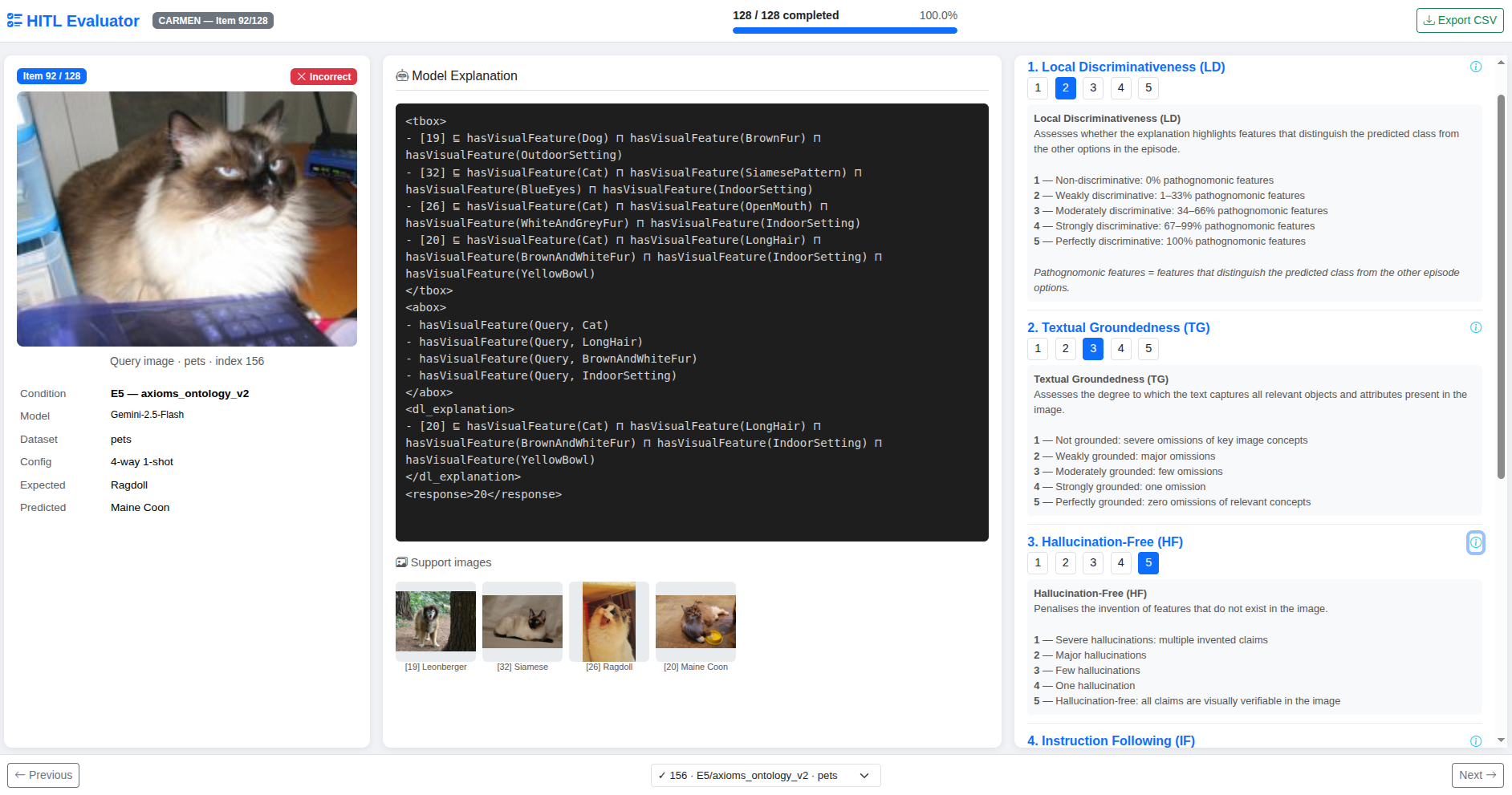}
\caption{Screenshot of the human annotation interface used for
  judge validation.
  \textbf{Left column:} query image with ground-truth class and
  predicted label; a green border indicates a correct prediction,
  red an incorrect one.
  \textbf{Centre column:} explanation generated by the classifier
  model (preserving XML tag formatting); below it, a collapsible
  panel displays the labelled support images with their class names.
  \textbf{Right column:} scoring form for the nine XAI metrics;
  each metric shows radio buttons for scores 1--5, with an
  expandable rubric (button \textbf{i}) providing the full scoring
  criteria; a progress bar at the top indicates how many of the
  128 assigned items have been completed.}
\label{fig:hitl_app}
\end{figure*}

% ---- H.2 ----
\subsection*{Sampling Protocol}

From the 4{,}608 trials with generated explanations (conditions
E2--E5), we selected 192 via stratified random sampling across
64~strata---unique combinations of explanation condition, model,
and dataset ($4 \times 4 \times 4$)---with 3~items per stratum.
This corresponds to 4.2\% of the total, a proportion consistent
with established practice for human validation of LLM-as-a-judge
pipelines~\citep{zheng2023judging, chiang2023large}.
Stratified sampling ensures uniform coverage of all experimental
combinations, which is methodologically superior to a larger
simple random sample that could under-represent rare or
difficult conditions~\citep{chiang2023large}.

Within each stratum, items were selected at random (seed\,=\,42)
with the constraint that at least one incorrectly classified trial
be included, to avoid positive-outcome bias --- models tend to
produce higher-quality explanations when the classification is
correct.
The 14~strata for which no incorrect predictions were available
(concentrated in Flowers and Pets, where model accuracy exceeds
99\%) were filled entirely with correct predictions;
these exceptions are documented in the sampling CSV provided in
the repository.

The 192~items were distributed among three researchers such that
each item was scored by exactly two independent annotators,
providing complete double-coverage of the sample.
Each researcher evaluated 128~items.
Due to sequential ordering of items by explanation condition prior
to block assignment, the distribution of conditions is not uniform
across annotator pairs; we therefore focus our analysis on
judge--human correlation rather than inter-annotator agreement,
which would conflate annotator calibration differences with
material difficulty.

% ---- H.3 ----
\subsection*{Evaluation Metrics}

We report two complementary statistics~\citep{mathur2020tangled} and an additional proposed metric to capture the singularities of the evaluation.

\textbf{Spearman rank correlation ($\rho$)} measures whether the
judge LLM ranks items from highest to lowest explanation quality
in the same order as the human annotators.
It is computed between the judge score and the mean of the two
human scores for each of the 192~items.
$\rho$ ranges from $-1$ to $1$: a value of $1$ indicates identical
ranking, $0$ indicates no relationship, and negative values
indicate opposite orderings.
Unlike Pearson correlation, Spearman makes no assumption of
linearity or normality, making it appropriate for ordinal 1--5
scales.
We adopt $\rho \geq 0.60$ as the threshold for substantial
agreement, following the scale of \citet{landis1977kappa} as
applied to rank-based measures in natural language generation
(NLG) evaluation \citep{zheng2023judging, liu2023g, chiang2023large}.

\textbf{Mean Absolute Error (MAE)} measures the average absolute
difference in points between the judge score and the human mean.
It ranges from $0$ (perfect agreement) to $4$ (maximum possible
error on a 1--5 scale).
MAE complements $\rho$ by detecting systematic magnitude biases:
a judge that consistently scores two points above or below humans
would yield a high MAE even if $\rho$ is high, because $\rho$
is sensitive only to ranking order, not to absolute values.

\textbf{Tolerance-Weighted Agreement (TWA)} To complement error magnitude (MAE) and rank correlation (Spearman $\rho$) evaluations, we introduce the Tolerance-Weighted Agreement (TWA) metric. Since qualitative 1-5 ordinal scales are inherently susceptible to subjective human calibration biases, a one-point divergence (e.g., a 4 versus a 5) often reflects natural annotator noise rather than structural semantic disagreement.  To mitigate this effect and capture the operational utility of the LLM-as-a-judge, TWA acts as a relaxed categorical accuracy that grants partial credit to adjacent predictions. 

%Agrego
Let $y_{human}^{(i)}, y_{judge}^{(i)} \in \{1, 2, 3, 4, 5\}$ represent the scores assigned by the human and the algorithmic judge for the $i$-th evaluated instance, respectively. Defining the absolute distance between the judge and human scores as $d_i = |y_{judge}^{(i)} - y_{human}^{(i)}|$, the item-level scoring function $S(d_i)$ is formulated as:$$S(d_i) = \begin{cases} 1 & \text{if } d_i = 0 \\ 0.75 & \text{if } d_i = 1 \\ 0 & \text{if } d_i > 1 \end{cases}$$The overall Tolerance-Weighted Agreement (TWA) is then calculated as the average score across all $N$ instances:$$TWA = \frac{1}{N} \sum_{i=1}^{N} S(d_i)$$

TWA provides a pragmatic assessment of functional alignment. A score of 0.75 was selected for instances with a distance of 1 ($d=1$) to provide a near-perfect weight, thereby rewarding adjacent predictions (buffering human subjectivity) while still distinguishing them from exact agreement. To ensure the metric remains rigorously intolerant of severe evaluative misalignments, any absolute distance strictly greater than one ($d > 1$) between the judge and human scores is penalized with a score of $0$. Eliminating partial credit for these severe cases prevents artificial inflation of the agreement score. This adjustment ensures that the TWA is both operationally pragmatic regarding minor human subjectivity and rigorously intolerant of severe evaluative misalignments.

%Let $y_{human}, y_{judge} \in \{1, 2, 3, 4, 5\}$ represent the scores assigned by the human and the algorithmic judge, respectively. Defining the absolute distance between the judge and human scores as 
%$d = |y_{judge} - y_{human}|$, the metric is formulated as:$$TWA(d) = \begin{cases} 1 & \text{if } d = 0 \\ 0.75 & \text{if } d = 1 \\ 0 & \text{if } d > 1 \end{cases}$$

%TWA provides a pragmatic assessment of functional alignment. By highly rewarding $d=1$ distances (buffering human subjectivity) and applying a strict penalty to larger divergences ($d>1$), this metric interpretably isolates the proportion of algorithmic verdicts that are practically equivalent to human judgments. A score of 0.75 was selected for instances with a distance of 1 ($d=1$) to provide a near-perfect weight, thereby rewarding adjacent predictions while still distinguishing them from exact agreement. Its inclusion alongside MAE and Spearman enables robust triangulation: TWA confirms practical, item-by-item consensus. 

%TWA score equal or greater than 0.75 indicates that the judge and human ratings differ by at most one point on average, with values approaching 1.0 reflecting exact agreement. Since most dimensions evaluated yielded scores at or above this threshold, the pragmatic discrepancies between the LLM and human annotators are largely confined to adjacent ratings rather than severe misalignments. 

TWA score equal or greater than 0.75 indicates that the judge and human ratings differ by at most one point on average, with values approaching 1.0 reflecting exact agreement. By combining this reward with a strict penalty for larger divergences ($d>1$), this metric interpretably isolates the proportion of algorithmic verdicts that are practically equivalent to human judgments. Its inclusion alongside MAE and Spearman enables robust triangulation: TWA confirms practical, item-by-item consensus.

% ---- H.4 ----
\subsection*{Results}

\begin{figure}[!t]
\centering
\includegraphics[width=\linewidth]{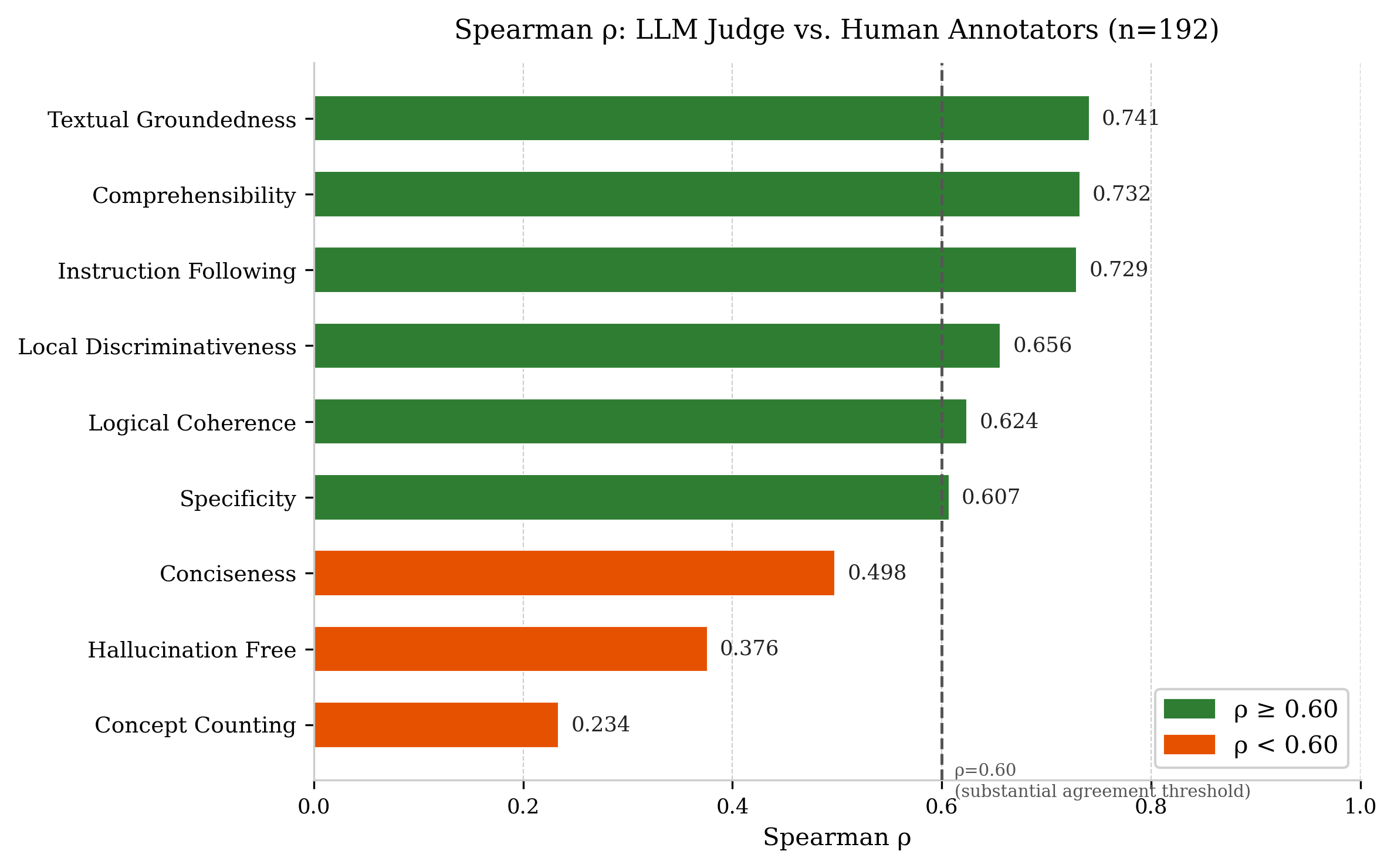}
\caption{Spearman $\rho$ between the LLM-as-a-judge scores and
  the mean of two independent human annotators, for each of the
  nine XAI metrics ($n = 192$ stratified trials).
  Each bar shows the $\rho$ value for one metric; bars are ordered
  from highest to lowest.
  \textbf{Green bars} ($\rho \geq 0.60$) indicate metrics for
  which the judge's ranking closely matches the human ranking,
  consistent with substantial agreement~\citep{landis1977kappa}.
  \textbf{Orange bars} ($\rho < 0.60$) indicate metrics where
  alignment is weaker.
  The dashed vertical line marks the $\rho = 0.60$ threshold.
  Six of nine metrics exceed this threshold, including Local
  Discriminativeness ($\rho = 0.656$), the metric most central
  to the paper's conclusions.}
\label{fig:hitl_rho}
\end{figure}

\begin{figure}[!t]
\centering
\includegraphics[width=\linewidth]{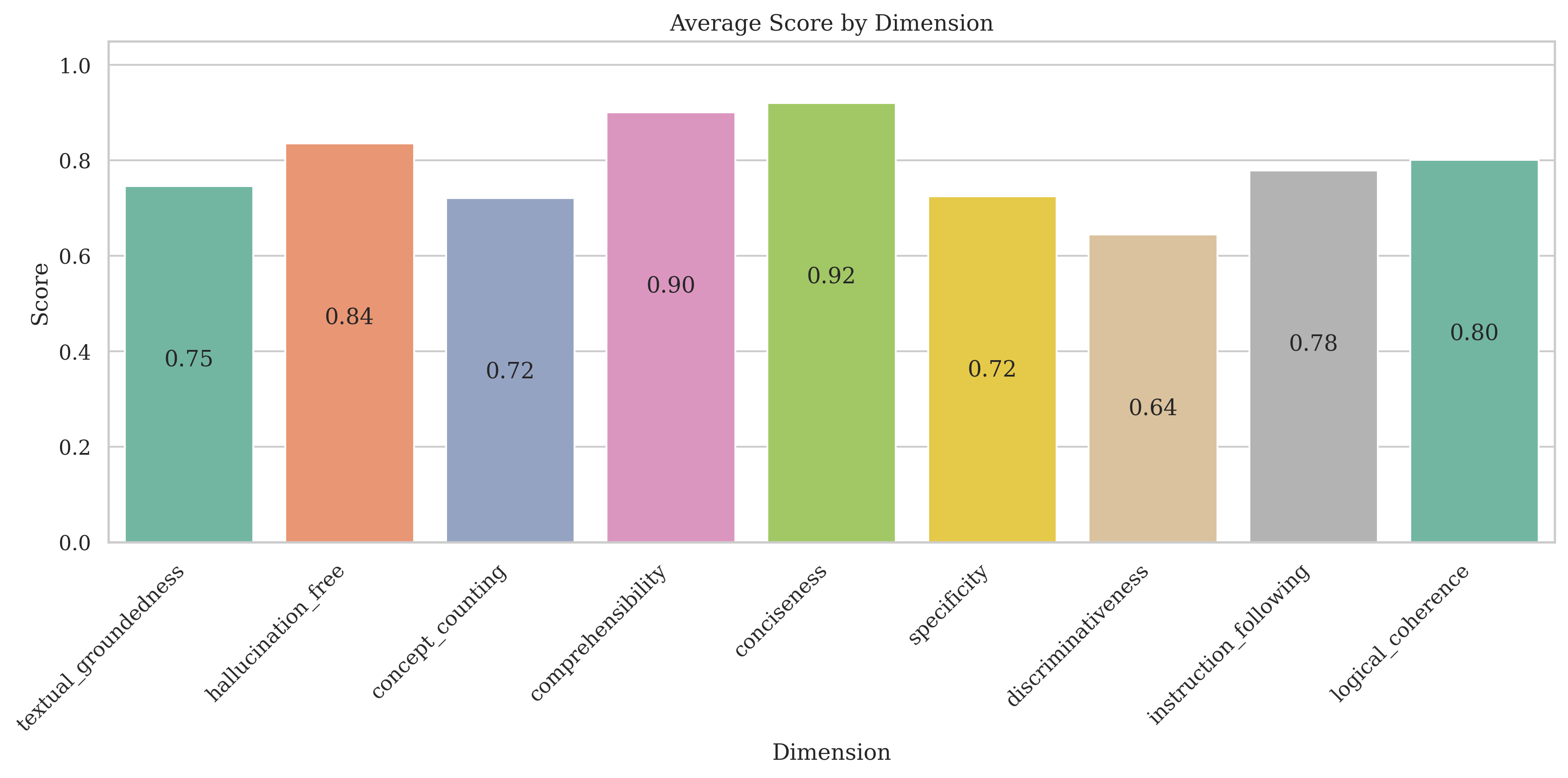}
\caption{Average TWA scores by XAI metric. The LLM-as-a-judge demonstrates strong practical alignment on dimensions like Conciseness (0.92) and Comprehensibility (0.90). Evaluative dimensions requiring class-contrastive logic, such as Local Discriminativeness (0.64), maintain robust but comparatively lower agreement.}
\label{fig:twa_dimension}
\end{figure}

\begin{figure}[H]
\centering
\includegraphics[width=\linewidth]{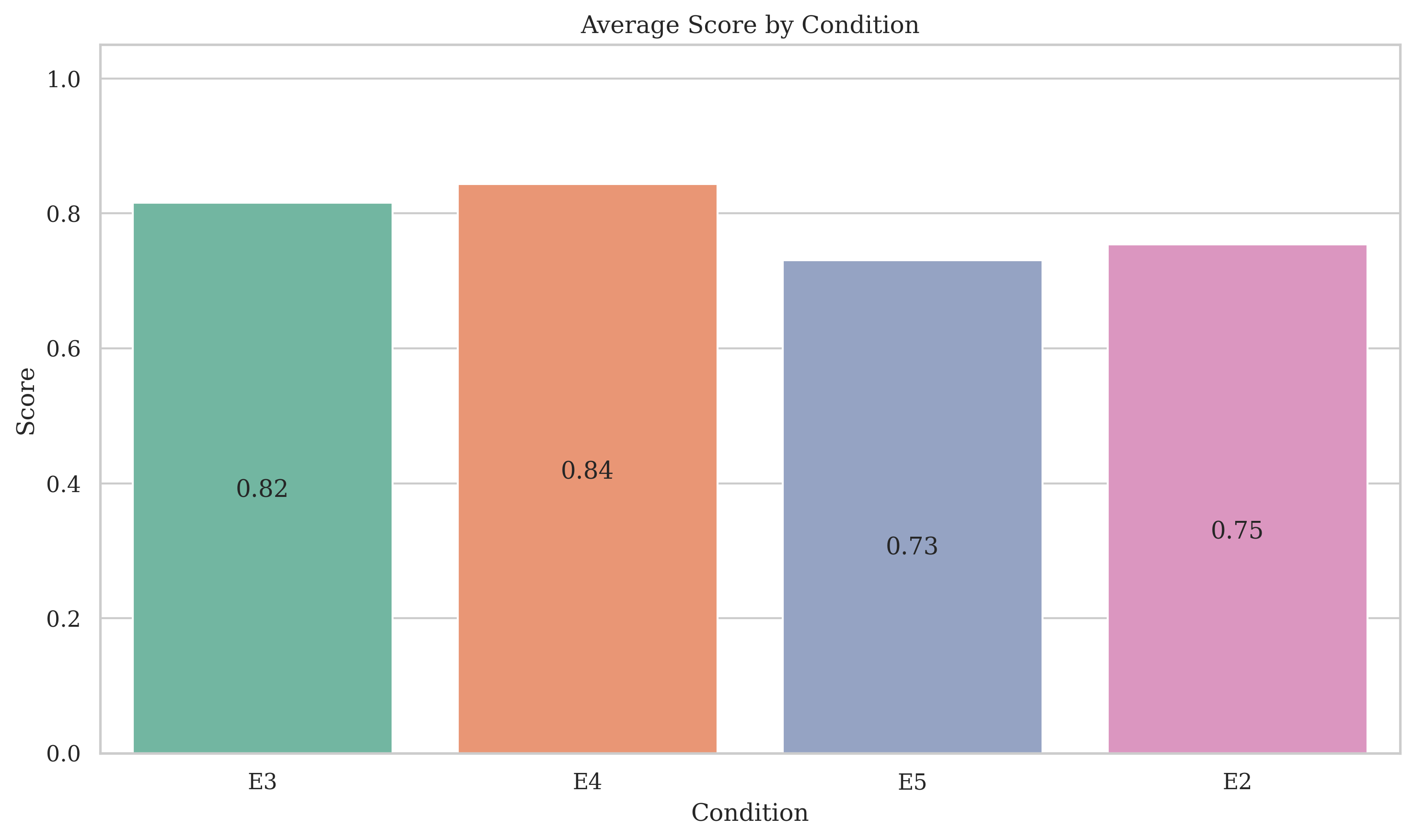}
\caption{Average TWA scores categorized by explanation condition. Structured intermediate formats like Feature-value pairs (E4, 0.84) and Features (E3, 0.82) yield the highest human-judge alignment, outperforming free-text NLE (E2, 0.75) and formal DL Axioms (E5, 0.73).}
\label{fig:twa_condition}
\end{figure}

\begin{figure}[!t]
\centering
\includegraphics[width=\linewidth]{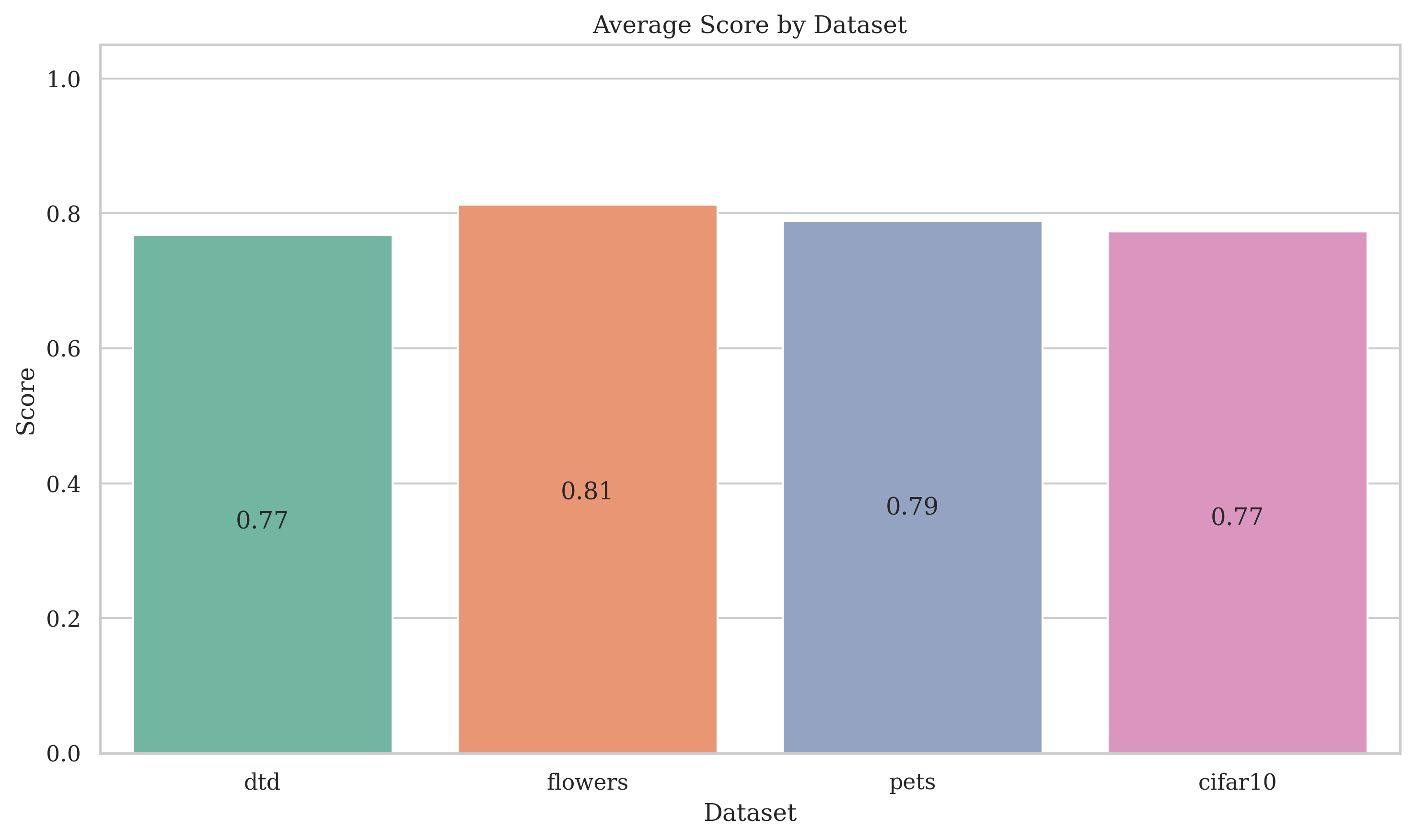}
\caption{Average Tolerance-
Weighted Agreement (TWA) scores across the four evaluated image datasets. Human-judge agreement remains highly consistent across different visual domains, ranging from 0.77 on abstract textures (DTD) and generic objects (CIFAR-10) to 0.81 on fine-grained categories (Flowers), indicating stability regardless of dataset complexity.}
\label{fig:twa_dataset}
\end{figure}

Table~\ref{tab:hitl_results} reports Spearman $\rho$ and MAE for
all nine metrics, and Figure~\ref{fig:hitl_rho} visualises the
$\rho$ values against the $\rho = 0.60$ threshold.
TWA scores are reported by XAI metric in Figure~\ref{fig:twa_dimension},
by explanation condition in Figure~\ref{fig:twa_condition},
and by dataset in Figure~\ref{fig:twa_dataset}.

\begin{table}[H]
\centering
\caption{Human validation results for the LLM-as-a-judge pipeline
  ($n = 192$ stratified trials, 3~annotators, each item scored by
  2~annotators independently).
  \textbf{Spearman $\rho$}: rank correlation between judge score
  and mean human score (range $-1$ to $1$; higher is better;
  $\rho \geq 0.60$ indicates substantial agreement).
  \textbf{MAE}: mean absolute error in points between judge score
  and mean human score (range $0$--$4$; lower is better).
  A horizontal rule separates metrics above and below the
  $\rho = 0.60$ threshold.}
\label{tab:hitl_results}
\small
\setlength{\tabcolsep}{6pt}
\renewcommand{\arraystretch}{1.15}
\begin{tabular}{@{}lcc@{}}
\toprule
\textbf{Explainability Metric} & \textbf{Spearman $\rho$} & \textbf{MAE} \\
\midrule
Textual Groundedness (TG)      & 0.741 & 0.672 \\
Comprehensibility (CP)         & 0.732 & 0.276 \\
Instruction Following (IF)     & 0.729 & 0.555 \\
Local Discriminativeness (LD)  & 0.656 & 0.865 \\
Logical Coherence (LC)         & 0.624 & 0.560 \\
Specificity (S)                & 0.607 & 0.646 \\
\midrule
Conciseness (Cn)               & 0.498 & 0.198 \\
Hallucination Free (HF)        & 0.376 & 0.474 \\
Concept Counting (CC)          & 0.234 & 0.867 \\
\bottomrule
\end{tabular}
\end{table}

% ---- H.5 ----
\subsection*{Discussion}

The judge achieves $\rho > 0.60$ on the six metrics most relevant
to the between-condition comparisons reported in the paper:
Textual Groundedness, Comprehensibility, Instruction Following,
Local Discriminativeness, Logical Coherence, and Specificity.
The lower $\rho$ values for Conciseness, Hallucination Free, and
Concept Counting do not reflect judge unreliability.
The first two suffer from a ceiling effect: most explanations
score near 5 on these dimensions across all conditions, so there
is little variation to rank.
Concept Counting is structurally different: it requires exact
enumeration of visual elements in the image, a task that proved
difficult to apply consistently even among human annotators.
The higher MAE values observed for Local Discriminativeness and
Textual Groundedness reflect annotator calibration differences
rather than systematic judge error: the judge's absolute scores
align closely with two of the three annotators, while the third
annotator applied consistently higher ratings across all metrics,
inflating the human mean for those dimensions.

To reconcile these statistical nuances, the Tolerance-Weighted Agreement (TWA) provides a clearer picture of the judge’s operational reliability. By mitigating the impact of the ceiling effects mentioned above and the calibration discrepancies—granting partial credit for adjacent scores that reflect subjective human noise while strictly penalizing severe semantic disagreements ($d > 1$) with a score of zero—the TWA confirms that the judge’s evaluations remain functionally aligned with human consensus. Ultimately, the consistently high TWA scores confirm that the LLM-as-a-judge succeeds as a robust and practical evaluator, even in dimensions where $\rho$ or MAE penalize strict numerical divergence.

\end{document}